\documentclass[runningheads]{llncs}
\usepackage{graphicx}
\usepackage{breakcites}

\usepackage{hyperref}
\usepackage{graphicx}
\usepackage{amsmath}
\usepackage{amssymb}
\usepackage{booktabs}
\usepackage{verbatim}
\usepackage{multirow}
\usepackage{graphicx}  
\usepackage{url}  
\usepackage{color}
\usepackage{colortbl}
\usepackage{comment}
\usepackage[rightcaption]{sidecap}
\usepackage{wrapfig}
\usepackage{subfigure}
\newcommand{\figref}[1]{Fig. \ref{#1}}

\usepackage{caption}
\usepackage{pifont}

\makeatletter
\def\hlinewd#1{%
\noalign{\ifnum0=`}\fi\hrule \@height #1 \futurelet
\reserved@a\@xhline}

\newcommand{\cmark}{\ding{51}}%
\newcommand{\xmark}{\ding{55}}%

\definecolor{recolor}{rgb}{0,1,0}

\definecolor{srcolor}{rgb}{1,0,0}
\newcommand{\sr}[1]{\textcolor{srcolor}{{#1}}}

\definecolor{stevecolor}{rgb}{0.7,0,0.7}

\definecolor{shcolor}{rgb}{0,0,1}

\usepackage[capitalize]{cleveref}
\crefname{section}{Sec.}{Secs.}
\Crefname{section}{Section}{Sections}
\Crefname{table}{Table}{Tables}
\crefname{table}{Tab.}{Tabs.}

\usepackage[accsupp]{axessibility}  

\begin{document}
\pagestyle{headings}
\mainmatter
\def\ECCVSubNumber{1987}  

\title{Cost Aggregation with 4D Convolutional Swin Transformer for Few-Shot Segmentation} 

\titlerunning{Cost Aggregation with 4D Convolutional Swin Transformer}
%
\author{Sunghwan Hong\inst{1,}$^\star$  \and
Seokju Cho\inst{1,}\thanks{Equal contribution}\and
Jisu Nam\inst{1}\and
Stephen Lin\inst{2}\and \\
Seungryong Kim\inst{1}} 
\authorrunning{S. Hong et al.}
%
\institute{Korea University, Seoul, Korea\\
\email{\{sung$\_$hwan,seokju$\_$cho,18wltnzzang,seungryong$\_$kim\}@korea.ac.kr}\and
Microsoft Research Asia, Beijing, China\\
\email{stevelin@microsoft.com }}

\maketitle

\begin{abstract}

This paper presents a novel cost aggregation network, called Volumetric Aggregation with Transformers (VAT), for few-shot segmentation. The use of transformers can benefit correlation map aggregation through self-attention over a global receptive field. However, the tokenization of a correlation map for transformer processing can be detrimental, because the discontinuity at token boundaries reduces the local context available near the token edges and decreases inductive bias. To address this problem, we propose a 4D Convolutional Swin Transformer, where a high-dimensional Swin Transformer is preceded by a series of small-kernel convolutions that impart local context to all pixels and introduce convolutional inductive bias. We additionally boost aggregation performance by applying transformers within a pyramidal structure, where aggregation at a coarser level guides aggregation at a finer level. 
Noise in the transformer output is then filtered in the subsequent decoder with the help of the query’s appearance embedding. With this model, a new state-of-the-art is set for all the standard benchmarks in few-shot segmentation. It is shown that VAT attains state-of-the-art performance for semantic correspondence as well, where cost aggregation also plays a central role. Code and trained models are available at~\url{https://seokju-cho.github.io/VAT/}.

\end{abstract}

\section{Introduction}
Semantic segmentation is a fundamental computer vision task that aims to label each pixel in an image with its corresponding class. Substantial progress has been made in this direction with the help of deep neural networks and large-scale datasets containing ground-truth segmentation annotations~\cite{long2015fully,noh2015learning,chen2017deeplab,chen2018encoder,tao2020hierarchical}. Manual labeling of pixel-wise segmentation maps, however, requires considerable labor, making it difficult to add new classes. Towards reducing reliance on labeled data, attention has increasingly focused on \textit{few}-shot segmentation~\cite{ravi2016optimization,shaban2017one}, where only a handful of support images and their associated masks are used in predicting the segmentation of a query image.

\begin{figure}[t]
\centering
\renewcommand{\thesubfigure}{}
	\subfigure[]
	{\includegraphics[width=1\linewidth]{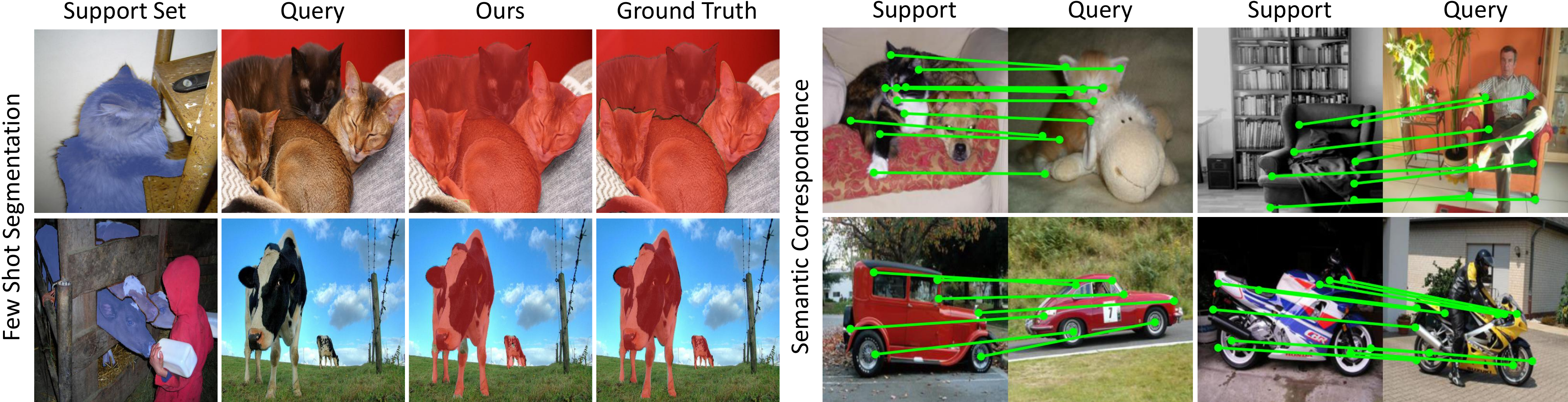}}\\
  \vspace{-15pt}
  \caption{\textbf{Our VAT reformulates few-shot segmentation as semantic correspondence.} VAT sets a new state-of-the-art in few-shot segmentation, and attains state-of-the-art performance for semantic correspondence as well.}
\label{fig1}\vspace{-10pt}
\end{figure}

The key to few-shot segmentation is in making effective use of the few support samples. Many works attempt this by extracting a prototype model from the samples and using it for feature comparison with the query~\cite{snell2017prototypical,dong2018few,liu2020part,yang2020prototype}. However, such approaches disregard pixel-level pairwise relationships between support and query features or the spatial structure of features, which may lead to sub-optimal results.
To account for such relationships, we observe that few-shot segmentation can be reformulated as semantic correspondence, which aims to find pixel-level correspondences across semantically similar images which may contain large intra-class appearance and geometric variations~\cite{ham2016proposal,ham2017proposal,min2019spair}. Recent semantic correspondence models~\cite{rocco2017convolutional,kim2017fcss,rocco2018end,Rocco18b,min2019hyperpixel,min2020learning,liu2020semantic,truong2020glu,min2021convolutional} follow the classical matching pipeline~\cite{scharstein2002taxonomy,philbin2007object} of feature extraction, cost aggregation and flow estimation. The cost aggregation stage, where matching scores are refined to produce more reliable correspondence estimates, is of particular importance and has been the focus of much research~\cite{Rocco18b,min2019hyperpixel,rocco2020efficient,jeon2020guided,liu2020semantic,li2020correspondence,min2021convolutional,cho2021semantic}. Recently, CATs~\cite{cho2021semantic} proposed to use vision transformers~\cite{dosovitskiy2020image} for cost aggregation, but its quadratic complexity to the number of input tokens limits its applicability. It also disregards the spatial structure of matching costs, which may hurt its performance.

In the area of few-shot segmentation, there also exist methods that attempt to leverage pairwise information by refining features through cross-attention~\cite{zhang2021few} or graph attention~\cite{zhang2019pyramid,wang2020few,xie2021scale}. However, they solely rely on raw correlation maps without aggregating the matching scores. As a result, their correspondence may suffer from ambiguities caused by repetitive patterns or background clutters~\cite{rocco2017convolutional,kim2017fcss,lee2019sfnet,truong2020glu,Hong_2021_ICCV}. 
To address this, HSNet~\cite{min2021hypercorrelation} aggregates the matching scores with 4D convolutions, but its limited receptive fields prevent long-range context aggregation and it lacks an ability to adapt to the input content due to the use of fixed kernels. 
In this paper, we introduce a novel cost aggregation network, called Volumetric Aggregation with Transformers (VAT), that tackles the few-shot segmentation task through a proposed 4D Convolutional Swin Transformer. Specifically, we first extend Swin Transformer~\cite{liu2021swin} and its patch embedding module to handle a high-dimensional correlation map. The patch embedding module is further extended by incorporating 4D convolutions that alleviate issues caused by patch embedding, \textit{i.e.,} limited local context near patch boundaries and low inductive bias. The high-dimensional patch embedding module is designed as a series of overlapping small-kernel convolutions, bringing local contextual information to each pixel and imparting convolutional inductive bias. To further boost performance, we compose our architecture with a pyramidal structure that takes the aggregated correlation maps at a coarser level as additional input at a finer level, providing hierarchical guidance. Our affinity-aware decoder then refines the aggregated matching scores in a manner that exploits the higher-resolution spatial structure given by the query’s appearance embedding and finally outputs the segmentation mask prediction. 

We demonstrate the effectiveness of our method on several benchmarks~\cite{shaban2017one,lin2014microsoft,li2020fss}. Our work attains state-of-the-art performance on all the benchmarks for few-shot segmentation and even for semantic correspondence, highlighting the importance of cost aggregation for both tasks and showing its potential for general matching. We also include ablation studies to justify our design choices.

\section{Related Work}
\subsubsection{Few-shot Segmentation.} Inspired by the few-shot learning paradigm~\cite{ravi2016optimization,snell2017prototypical}, which learns to learn a model for a novel task with only a limited number of samples, few-shot segmentation has received considerable attention. Following the success of~\cite{shaban2017one}, prototypical networks~\cite{snell2017prototypical} and numerous other works~\cite{dong2018few,nguyen2019feature,siam2019adaptive,wang2019panet,liu2020part,yang2020prototype,liu2020crnet,xie2021few,yang2021mining,sun2021boosting,zhang2021prototypical,li2021adaptive} proposed to extract
a prototype from support samples, which is used to identify foreground features in the query.
In addition, inspired by~\cite{zhang2019canet} which observed that simply adding high-level features in feature processing leads to a performance drop,~\cite{tian2020prior} proposed to instead utilize high-level features to compute a prior map that helps to identify targets in the query image. Many variants~\cite{sun2021boosting,zhang2021self} extended this idea of utilizing prior maps to act as additional information for aggregating feature maps.

However, as methods based on prototypes or prior maps have apparent limitations, \textit{e.g.}, disregarding pairwise relationships between support and query features or spatial structure of feature maps, numerous recent works~\cite{zhang2019pyramid,wang2020few,min2021hypercorrelation,xie2021scale,liu2021few} utilize a correlation map to leverage the pairwise relationships between source and query features. Specifically,~\cite{zhang2019pyramid,wang2020few,xie2021scale} use graph attention, HSNet~\cite{min2021hypercorrelation} proposes 4D convolutions to exploit multi-level features, and~\cite{liu2021few} formulates the task as an optimal transport problem. However, these approaches do not provide a means to aggregate the matching scores, solely utilize convolutions for cost aggregation, or use a handcrafted method that is neither learnable nor robust to severe deformations. 

Recently,~\cite{zhang2021few} utilized transformers and proposed to use a cycle-consistent attention mechanism to refine the feature maps to become more discriminative, without considering aggregation of matching scores.~\cite{sun2021boosting} propose a global and local enhancement module to refine the features using transformers and convolutions, respectively.~\cite{lu2021simpler} focuses solely on the transformer-based classifier by freezing the encoder and decoder. Unlike these works, we propose a 4D Convolutional Swin Transformer for an enhanced and efficient cost aggregation.  \vspace{-10pt}  

\subsubsection{Semantic Correspondence.}
The objective of semantic correspondence is to find correspondences between semantically similar images with additional challenges posed by large intra-class appearance and geometric variations~\cite{liu2020semantic,cho2021semantic,min2021convolutional}. This is highly similar to the few-shot segmentation setting in that few-shot segmentation also aims to label objects of the same class with large intra-class variation, and thus recent works on both tasks have taken similar approaches. The latest methods~\cite{Rocco18b,min2019hyperpixel,rocco2020efficient,jeon2020guided,liu2020semantic,li2020correspondence,min2021convolutional,cho2021semantic} in semantic correspondence focus on the cost aggregation stage to find reliable correspondences and demonstrated its importance. Among them,~\cite{min2021convolutional} proposed to use 4D convolutions for cost aggregation, though exhibiting apparent limitations due to the limited receptive fields of convolutions and lack of adaptability. CATs~\cite{cho2021semantic} resolves this issue and sets a new state-of-the-art by leveraging transformers~\cite{vaswani2017attention} to aggregate the cost volume. However, it disregards the spatial structure of correlation maps and imparts less inductive bias, \textit{i.e.,} translation equivariance, which limits its generalization power~\cite{liu2021swin,dai2021coatnet,d2021convit}. Moreover, its quadratic complexity may limit applicability when it is used to aggregate correlation maps on its own. In this paper, we propose to resolve the aforementioned issues. \vspace{-10pt}


\subsubsection{Vision Transformer.}
Recently, transformer~\cite{vaswani2017attention}, the standard architecture in Natural Language Processing (NLP), has been widely adopted in Computer Vision. Since the pioneering work on ViT~\cite{dosovitskiy2020image}, numerous works~\cite{lu2021simpler,zhang2021few,sun2021boosting,jiang2021cotr,wu2021fully,cho2021semantic,liu2021swin} have adopted transformers to replace CNNs or to be used together with CNNs in a hybrid manner. However, due to quadratic complexity to sequence length, transformers often suffer from large a computational burden. Efficient transformers~\cite{wang2020linformer,katharopoulos2020transformers,xiong2021nystromformer,wu2021fastformer} aim to reduce the computational load via an approximated or simplified self-attention. Swin Transformer~\cite{liu2021swin}, a network we extend from, reduces computation by performing self-attention within pre-defined local windows. However, these works inherit the issues caused by patch embedding, which we alleviate by incorporating 4D convolutions.     

\section{Methodology}
\subsection{Problem Formulation}
The goal of \textit{few}-shot segmentation is to segment objects from unseen classes in a query image given only a few annotated examples~\cite{vinyals2016matching}. To mitigate the overfitting caused by insufficient training data, we follow the common protocol of \textit{episodic} training~\cite{vinyals2016matching}.
Let us denote the training and test sets as $\mathcal{D}_\mathrm{train}$ and $\mathcal{D}_\mathrm{test}$, respectively, where the object classes of both sets do not overlap. Under the $K$-shot setting, multiple \textit{episodes} are formed from both sets, each consisting of a support set $\mathcal{S} = \{(x_{s}^{k}, m_{s}^{k})\}_{k=1}^{\mathrm{K}}$, where $(x_{s}^{k}, m_{s}^{k})$ is $k$-th support image and its corresponding mask pair, and a query sample $(x_q, m_q)$, where $x_q$ 
is a query image and $m_q$ is its paired mask. During training, our model takes a sampled episode from $\mathcal{D}_\mathrm{train}$ and learns a mapping from $\mathcal{S}$ and $x_q$ to a prediction $m_q$. At inference, our model predicts $\hat{m}_q$ given randomly sampled $\mathcal{S}$ and $x_q$ from $\mathcal{D}_\mathrm{test}$.
\begin{figure*}[t]
    \centering
    \renewcommand{\thesubfigure}{}
	\subfigure[]
	{\includegraphics[width=1\linewidth]{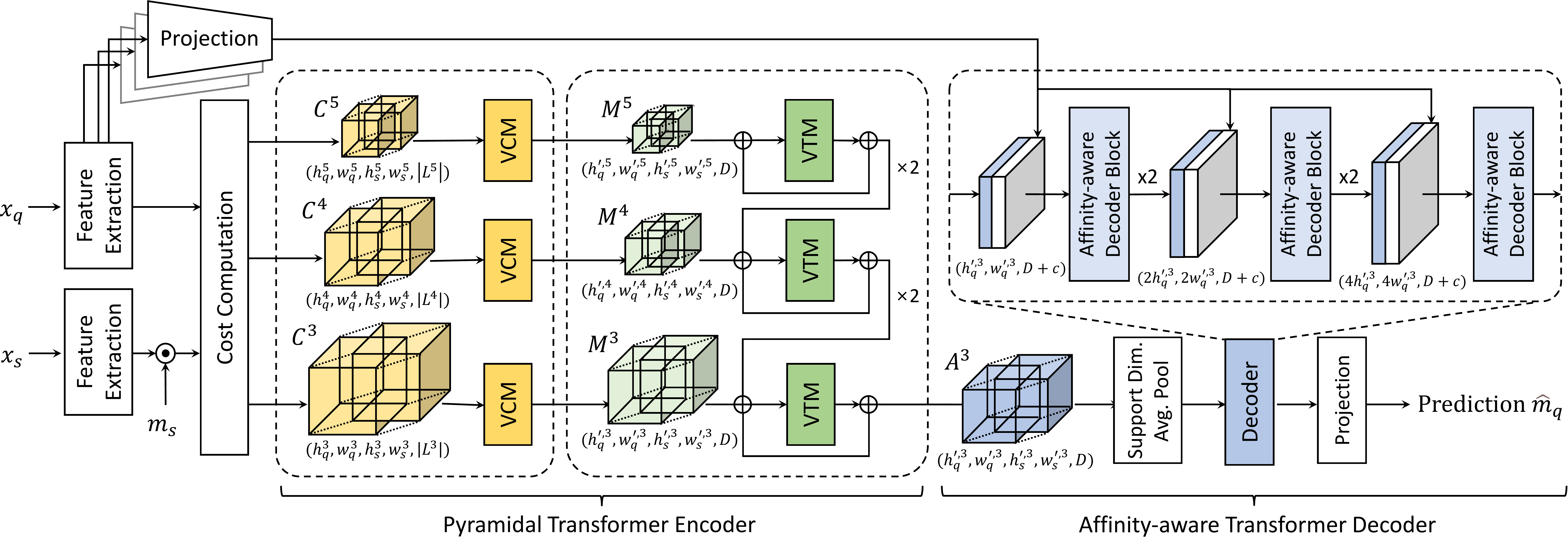}}\\ \vspace{-15pt}
    \caption{\textbf{Overall network architecture.} Our network consists of feature extraction and cost computation, a pyramidal transformer encoder, and an affinity-aware transformer decoder.} 
    \label{fig:overall}\vspace{-10pt}
\end{figure*}

\subsection{Motivation and Overview}
The key to few-shot segmentation is how to effectively utilize the support samples provided for a query image. While conventional methods~\cite{tian2020prior,sun2021boosting,zhang2021few,yang2021mining,li2021adaptive} utilize global- or part-level prototypes extracted from support features, recent methods~\cite{zhang2019pyramid,wang2020few,min2021hypercorrelation,xie2021scale,liu2021few,zhang2021few} instead leverage pairwise matching relationships between query and support. However, exploring such relationships is notoriously challenging due to intra-class variations, background clutters, and repetitive patterns. One of the state-of-the-art methods, HSNet~\cite{min2021hypercorrelation}, aggregates the matching scores with 4D convolutions. However, solely utilizing convolutions may limit performance due to limited receptive fields or lack of adaptability for convolutional kernels. While there has been no approach to aggregate the matching scores with transformers in few-shot segmentation, CATs~\cite{cho2021semantic} proposes cost aggregation with transformers in semantic correspondence, demonstrating the effectiveness of transformers as a cost aggregator. On the other hand, the quadratic complexity of transformers with respect to the number of tokens may limit its utility for segmentation. The absence of operations that impart inductive bias, \textit{i.e.,} translation equivariance, may limit its performance as well. 
Also, CATs~\cite{cho2021semantic} defines the tokens of a correlation map in a way that disregards spatial structure, which is likely to be harmful.

The proposed Volumetric Aggregation with Transformers (VAT) is designed to overcome these problems. In the following, we first describe its feature extraction and cost computation. We then present a general extension of Swin Transformer~\cite{liu2021swin} for cost aggregation. Subsequently, we present 4D Convolutional Swin Transformer for resolving the aforementioned issues. Lastly, we introduce several additional techniques including Guided Pyramidal Processing (GPP) and Affinity-aware Transformer Decoder (ATD) to further boost performance, and combine them to complete the design.

\subsection{Feature Extraction and Cost Computation}\label{sec:3.3}
We extract features from query and support images and compute an initial cost between them following the conventional process~\cite{rocco2017convolutional,sun2018pwc,Rocco18b,rocco2020efficient,truong2020glu,Hong_2021_ICCV,cho2021semantic}. Given query and support images, $x_q$ and $x_s$, we use a CNN~\cite{he2016deep,simonyan2014very} to produce a sequence of $L$ feature maps, $\{(F_{q}^{l},F_{s}^{l})\}^L_{l=1}$, where $F_{q}^{l}$ and $F_{s}^{l}$ denote query and support feature maps at the $l$-th level. A support mask, $m_s$, is used to encode segmentation information and filter out the background information as done in~\cite{li2021adaptive,min2021hypercorrelation,zhang2021self}. We obtain a masked support feature as $\hat{F}^{l}_{s} = {F}^{l}_{s} \ \odot \ \psi^{l}{(m_{s})}$,
where $\odot$ denotes the Hadamard product and $\psi^{l}(\cdot)$ denotes a function that resizes the given tensor followed by expansion along the channel dimension of the $l$-th layer.   

Given a pair of feature maps, $F^l_{q}$ and $F^l_{s}$, we compute a correlation map using the inner product between $l$-2 normalized features such that
\begin{equation}
    \mathcal{C}^l(i,j)=\mathrm{ReLU}\left(
    \frac{F^l_{q}(i)\cdot \hat{F}^l_{s}(j)}{\|F^l_{q}(i)\|\|\hat{F}^l_{s}(j)\|}\right),
\end{equation}
where $i$ and $j$ denote 2D spatial positions of feature maps. As done in~\cite{min2021hypercorrelation}, we collect correlation maps computed from all the intermediate features of the same spatial size and stack them to obtain a stacked correlation map $\mathcal{C}^p \in \mathbb{R}^{h_{q} \times w_{q} \times h_{s} \times w_{s} \times |\mathcal{L}^p|}$, where $(h_{q}, w_{q})$ and $(h_{s},  w_{s})$ are the height and width of the query and support feature maps, respectively, and $\mathcal{L}^p$ is a subset of CNN layer indices $\{1,...,L\}$ at pyramid layer $p$, containing correlation maps of identical spatial size. 

\subsection{Pyramidal Transformer Encoder}
In this section, we present 4D Convolutional Swin Transformer for aggregating the correlation maps and then incorporate it into a pyramidal architecture.  \vspace{-10pt}

\subsubsection{Cost Aggregation with Transformers.}
For a transformer to process a correlation map, a means for token reduction is essential, since it would be infeasible for even an efficient transformer~\cite{wang2020linformer,katharopoulos2020transformers,xiong2021nystromformer,wu2021fastformer,liu2021swin} to handle a correlation map otherwise. However, when one employs a transformer for cost aggregation, the problem of how to define the tokens for correlation maps, which differ in shape from images, text or features~\cite{vaswani2017attention,dosovitskiy2020image}, is non-trivial. 
The first attempt to process correlation maps is CATs~\cite{cho2021semantic}, which reshapes the 4D correlation maps into 2D maps and performs self-attention in 2D. This disregards the spatial structure of correlation maps, \textit{i.e.,} over both support and query, which could limit its performance. To address this, one may treat all the spatial entries, \textit{e.g.,} $h_q\times w_q \times h_s\times w_s$, as tokens and treat $\mathcal{L}^p$ as the feature dimension for tokens. However, this results in a substantial computational burden that increases with larger correlation maps. This prevents the use of standard transformers~\cite{vaswani2017attention,dosovitskiy2020image} and encourages use of efficient versions as in~\cite{wang2020linformer,katharopoulos2020transformers,xiong2021nystromformer,wu2021fastformer,liu2021swin}. However, the use of simplified (or approximated) self-attention may be sub-optimal for performance, as will be discussed in Section~\ref{sec:4.4}. Furthermore, as proven in the optical flow and semantic correspondence literature~\cite{sun2018pwc,rocco2020efficient}, neighboring pixels tend to have similar correspondences. To preserve the spatial structure of correlation maps, we choose to use Swin Transformer~\cite{liu2021swin} as it not only provides efficient self-attention computation, but also maintains the smoothness property of correlation maps while still providing sufficient long-range self-attention. 

\begin{SCfigure}[][t]
	\includegraphics[width=0.55\linewidth]{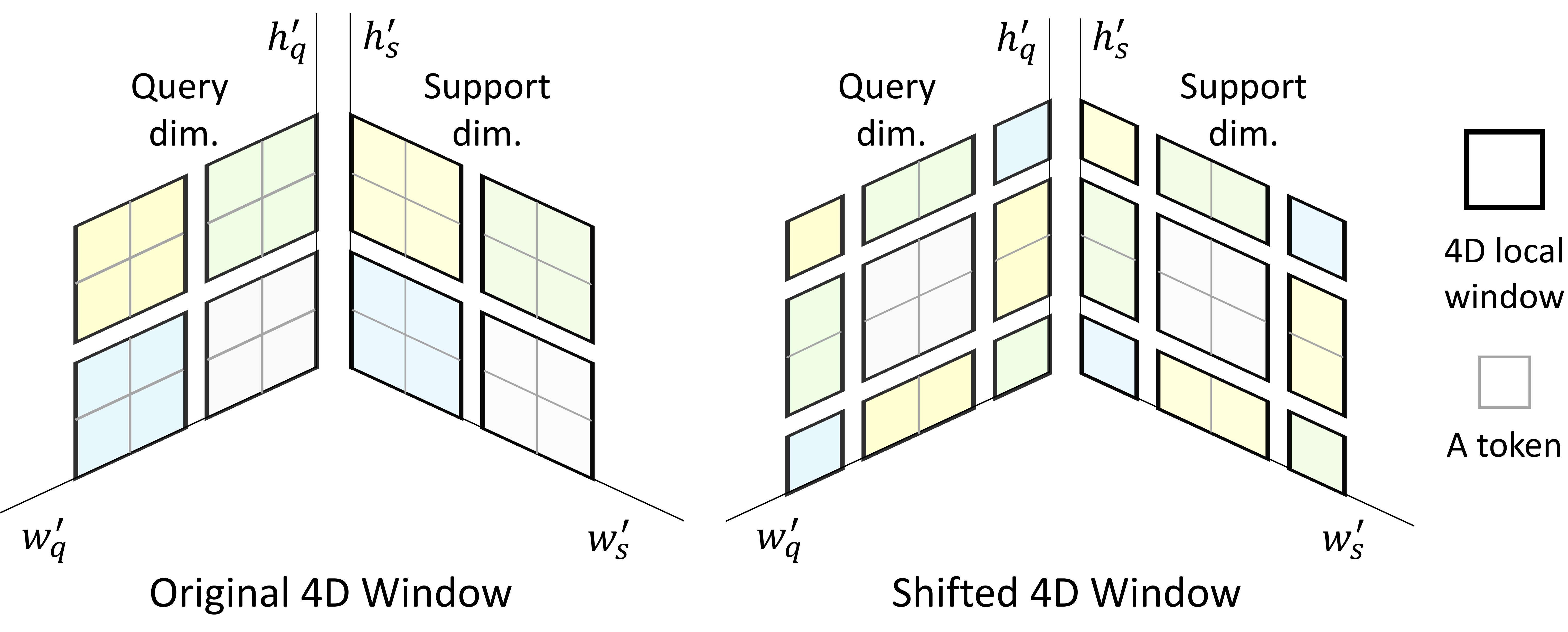}
	\caption{\textbf{Illustration of shifted 4D windows in VTM.} It computes self-attention within the partitioned windows and considers inter-window interactions by shifting the windows.}
	\label{fig:VTM}\vspace{-10pt}
\end{SCfigure}
To employ Swin Transformer~\cite{liu2021swin} for cost aggregation, we need to extend it to process higher dimensional input, specifically a 4D correlation map. We first follow the conventional patch embedding procedure~\cite{dosovitskiy2020image} to embed correlation maps, as they cannot be processed by transformers due to the large number of tokens. However, we extend the patch embedding module to a Volumetric Embedding Module (VEM) which handles higher dimensional inputs, such that $\mathcal{M}^p = \mathrm{VEM}(\mathcal{C}^p)$. Following a procedure similar to patch embedding, we reshape the correlation map to a sequence of flattened 4D windows using a large convolutional kernel, \textit{e.g.,} 16$\times$16$\times$16$\times$16. Then, we extend the self-attention computations, as shown in Fig.~\ref{fig:VTM}, by evenly partitioning the query and support spatial dimensions of $\mathcal{M}^p$ into non-overlapping sub-correlation maps $\mathcal{M}'^{,p} \in \mathbb{R}^{n\times n \times n \times n \times D}$. We compute self-attention within each partitioned sub-correlation map. Subsequently, we shift the windows by a displacement of $\left(\lfloor \frac{n}{2} \rfloor, \lfloor \frac{n}{2} \rfloor, \lfloor \frac{n}{2} \rfloor, \lfloor \frac{n}{2} \rfloor\right)$ pixels from the previously partitioned windows, then perform self-attention within the newly created windows. Then as done in the original Swin Transformer~\cite{liu2021swin}, we simply roll the correlation map back to its original form. In computing self-attention, we use relative position bias and take the values from an expanded parameterized bias matrix, following~\cite{hu2018relation,hu2019local,liu2021swin}. We leave the other components of Swin Transformer blocks unchanged, e.g., Layer Normalization (LN)~\cite{ba2016layer} and MLP layers. We call this extension the Volumetric Transformer Module (VTM). To summarize, the overall process is defined as:
\begin{equation}
  \mathcal{A}^p = \mathrm{VTM}(\mathcal{M}^p). 
  \vspace{-5pt}
\end{equation}

\subsubsection{4D Convolutional Swin Transformer.}
Although the proposed cost aggregation with transformers can solve the aforementioned issues of using CNNs and the high computational burden of using standard transformers, it may not avoid the issue that other transformers share~\cite{dosovitskiy2020image,wang2020linformer,katharopoulos2020transformers,xiong2021nystromformer,wu2021fastformer}: lack of translation equivariance. This is primarily caused by utilizing non-overlapping operations prior to self-attention computation. Although Swin Transformer alleviates the issue to some extent by using relative positioning bias~\cite{liu2021swin}, it provides an insufficient approximation. We argue that the Volumetric Embedding Module is what needs to be addressed as it leads to several issues. First, the use of large non-overlapping convolution kernels only provides limited inductive bias. Relatively lower translation equivariance is achieved from non-overlapping operations compared to that which are overlapping. This limited inductive bias results in relatively lower generalization power and performance~\cite{xiao2021early,d2021convit,dai2021coatnet,liu2021swin}. Furthermore, we argue that for dense prediction tasks, disregarding window boundaries due to non-overlapping kernels hurts overall performance due to discontinuity. 

\begin{figure*}[t]
    \centering
    \renewcommand{\thesubfigure}{}
	\subfigure[]
	{\includegraphics[width=1\linewidth]{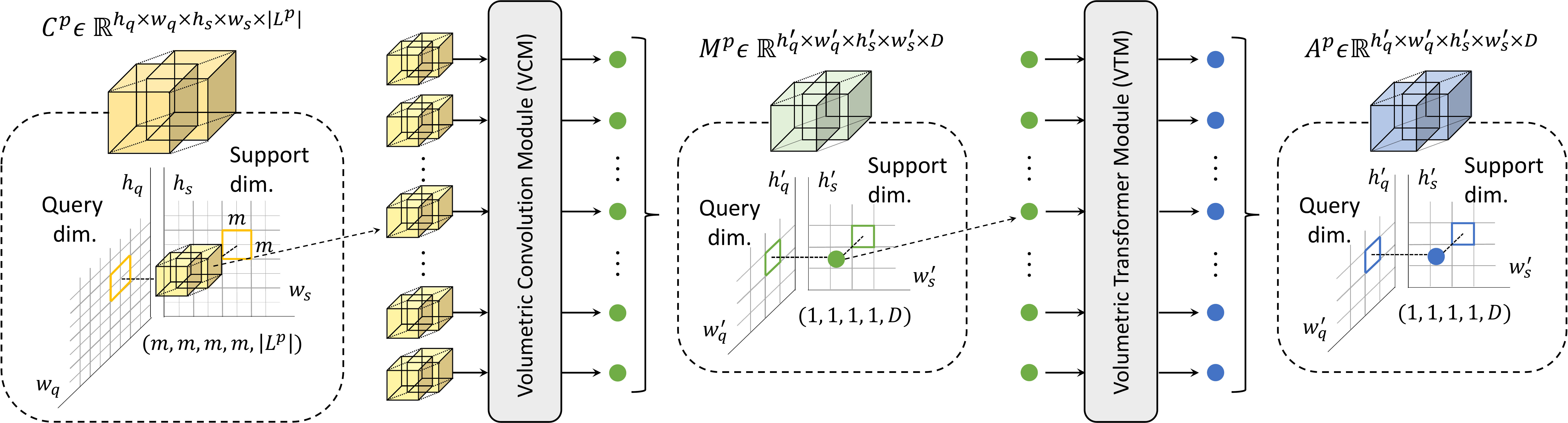}}\\\vspace{-10pt}
    \caption{\textbf{Overview of 4D Convolutional Swin Transformer.} We  replace the VEM with VCM and the output undergoes VTM for cost aggregation.}
    \label{fig:vem}\vspace{-10pt}
\end{figure*}

To address the above issues, we replace the Volumetric Embedding Module (VEM) with a module consisting of a series of overlapping convolutions, which we call the Volumetric Convolution Module (VCM). Concretely, we sequentially reduce spatial dimensions of the support and query by applying 4D spatial max-pooling, overlapping 4D convolutions, ReLU, and Group Normalization (GN), where we project the multi-level similarity vector at each 4D position, i.e., projecting a vector size of $|\mathcal{L}^p|$, to an arbitrary fixed dimension denoted as $D$. Considering receptive fields as a 4D window, i.e., $m\times m\times m\times m$, we obtain a tensor $\mathcal{C}^p \in \mathbb{R}^{{h}'^{,p}_{q}\times {w}'^{,p}_{q} \times {h}'^{,p}_{s} \times {w}'^{,p}_{s} \times D}$ from $\mathcal{C}^p$, where ${h}'^{,p}_{s}, {w}'^{,p}_{s}$, ${h}'^{,p}_{q}$, and ${w}'^{,p}_{q}$ are the processed sizes. Note that a different size of $m$ can be chosen for the support and query spatial dimensions. An overview of VCM is illustrated in Fig.~\ref{fig:vem}. 
Overall, we define such a process as the following:
\begin{equation}
\begin{split}
    \mathcal{M}^p = \mathrm{VCM}(\mathcal{C}^p).
\end{split}
\end{equation}
In this way, our model benefits from additional inductive bias as well as better handling at window boundaries. 

Moreover, to stabilize the learning, we propose an additional technique to enforce the networks to estimate residual matching scores as complementary details. We add residual connections in order to expedite the learning process~\cite{he2016deep,cho2021semantic,zhao2021multi}, accounting for the fact that at the initial phase when the input $\mathcal{M}^p$ is fed, erroneous matching scores are inferred due to randomly-initialized parameters of transformers, which could complicate the learning process as the networks need to learn the complete matching details from random matching scores. \vspace{-10pt}    

\subsubsection{Guided Pyramidal Processing.}
Following~\cite{min2021hypercorrelation,sun2021boosting}, we also employ a coarse-to-fine approach through pyramidal processing as illustrated in~\figref{fig:overall}. Motivated by numerous recent works~\cite{zhang2021few,min2021convolutional,cho2021semantic,min2021hypercorrelation} in both semantic matching and few-shot segmentation which have demonstrated that leveraging multi-level features can boost performance by a large margin, we also use a pyramidal architecture.

In our coarse-to-fine approach, which we refer to as Guided Pyramidal Processing (GPP), the aggregation of a finer-level correlation map $\mathcal{A}^p$ is guided by the aggregated correlation map of the previous (coarser) level $\mathcal{A}^{p+1}$. Concretely, an aggregated correlation map $\mathcal{A}^{p+1}$ is up-sampled into a map $\mathrm{up}(\mathcal{A}^{p+1})$ which is added to the next level's correlation map $\mathcal{A}^p$ to serve as guidance. This process is repeated until the finest-level aggregated map is computed and passed to the decoder. As shown in Table~\ref{tab:vatable}, GPP leads to appreciable performance gains.

With GPP, the pyramidal transformer encoder is finally defined as:
\begin{equation}
    \mathcal{A}^p =
    \mathrm{VTM}(\mathrm{VCM}(\mathcal{C}^p)+\mathrm{up}(\mathcal{A}^{p+1})),
\end{equation}
where $\mathrm{up}(\cdot)$ denotes bilinear upsampling. 

\subsection{Affinity-Aware Transformer Decoder}
Given the aggregated correlation map produced by the pyramidal transformer encoder, a transformer-based decoder generates the final segmentation mask. To improve performance, we propose to conduct further aggregation within the decoder with the aid of the appearance embedding obtained from query feature maps.
The query's appearance embedding can help in two ways. First, appearance affinity information is an effective guide for filtering noise in matching scores, as proven in the stereo matching literature, e.g., Cost Volume Filtering (CVF)~\cite{hosni2012fast,sun2018pwc}. In addition, the higher-resolution spatial structure provided by an appearance embedding can be exploited to improve up-sampling quality, resulting in a highly accurate prediction mask $\hat{m}^q$ where fine details are preserved.

For the design of our Affinity-aware Transformer Decoder (ATD), we take the average over the support image dimensions of $\mathcal{A}^p$, concatenate it with the appearance embedding from query feature maps, and then aggregate by  transformers~\cite{vaswani2017attention,wang2020linformer,wu2021fastformer,liu2021swin} with subsequent bilinear interpolation. 
The process is defined as the following:
\begin{equation}\label{eq:5}
\hat{m}_q = \mathrm{ATD}([ \mathcal{A}'^{,p},\mathcal{P}(F_{q})]),
\end{equation}
where $\mathcal{A}'^{,p} \in \mathbb{R}^{{h}'^{,p}_{q}\times {w}'^{,p}_{q} \times D}$ is extracted by average pooling on $\mathcal{A}^p$ over the spatial dimensions of the support image, $\mathcal{P}(\cdot)$ is a linear projection, $\mathcal{P}(F_{q})\in \mathbb{R}^{{h}'^{,p}_{q}\times {w}'^{,p}_{q} \times c}$, and $[\,\cdot,\cdot\,]$ denotes concatenation. We sequentially refine the output immediately after bilinear upsampling to recapture fine details and integrate appearance information.

\subsection{Extension to $K$-Shot Setting}
Given $K$ pairs of support image and mask $\{(x_{s}^{i}, m_{s}^{i})\}_{i=1}^{{K}}$ and a query image $x_q$, our model forward-passes $K$ times to obtain $K$ different query masks $\hat{m}^k_{q}$. We sum up all the $K$ predictions at each spatial location, and if the sum divided by $K$ exceeds a threshold $\tau$, the location is predicted as foreground, and otherwise it is background.

\begin{table}[t]
    \begin{center}
    \scalebox{0.65}{
    \begin{tabular}{cl|ccccccc|ccccccc|c}
            \toprule
            \multirow{2}{*}{\shortstack{Backbone\\network}} & \multirow{2}{*}{Methods} & \multicolumn{7}{c|}{1-shot} & \multicolumn{7}{c|}{5-shot} & \# learnable \\ 
            
            & & $5^{0}$ & $5^{1}$ & $5^{2}$ & $5^{3}$ &mIoU & FB-IoU & mBA & $5^{0}$ & $5^{1}$ & $5^{2}$ & $5^{3}$ &mIoU & FB-IoU & mBA & params \\
            \midrule
            

            
            
            \multirow{8}{*}{ResNet50~\cite{he2016deep}} & PANet~\cite{wang2019panet}       & 44.0 & 57.5 & 50.8 & 44.0  &49.1 &-& - & 55.3 & 67.2 & 61.3 & 53.2  &59.3 & - & -& 23.5M \\

            & PFENet~\cite{tian2020prior}  & 61.7 & 69.5 & 55.4 & 56.3 &60.8 & 73.3 &- & 63.1 & 70.7 & 55.8 & 57.9 &61.9 & 73.9 & -& 10.8M \\ 
            
            & ASGNet~\cite{li2021adaptive} &58.8 &67.9&56.8  &53.7  &59.3 &69.2 &- &63.4 &70.6 &64.2 &57.4&63.9 &74.2 &- &10.4M \\
            
            & CWT~\cite{lu2021simpler} &56.3 &62.0 &59.9  &47.2  &56.4 &- &- &61.3 &68.5 &68.5 &56.6 &63.7  &-& - &- \\
            & RePRI~\cite{boudiaf2021few}  & 59.8 & 68.3 & \underline{62.1} & 48.5  &59.7 & - &49.0 & 64.6 & 71.4 & \textbf{71.1} & 59.3 &66.6 & -&43.8  & - \\ 
            
            & HSNet~\cite{min2021hypercorrelation}   & 64.3 & 70.7 & 60.3 & \textbf{60.5} &\underline{64.0} & \underline{76.7}&\underline{53.9}  & 70.3 & \underline{73.2} & 67.4 & \textbf{67.1} &\underline{69.5} & \underline{80.6}&\underline{54.5} & \textbf{2.6M} \\
            & CyCTR~\cite{zhang2021few} &\underline{65.7} &\underline{71.0} &59.5  &59.7 &\underline{64.0} &-&- &\underline{69.3} &\underline{73.5} &63.8 &63.5  &67.5  &-&- &- \\\cline{2-17} \\[-2.0ex]
            & VAT (ours)& \textbf{67.6} & \textbf{72.0} &\textbf{62.3}  & \underline{60.1}& \textbf{65.5}& \textbf{77.8} & \textbf{54.4}
& \textbf{72.4} &\textbf{73.6} &\underline{68.6} &\underline{65.7} & \textbf{70.1} &\textbf{80.9} &\textbf{54.8} &\underline{3.2M} \\

            \midrule
            
            \multirow{9}{*}{ResNet101~\cite{he2016deep}} & FWB~\cite{nguyen2019feature}        & 51.3 & 64.5 & 56.7 & 52.2 &56.2 & - & -& 54.8 & 67.4 & 62.2 & 55.3 &59.9 & - &- & 43.0M \\

            & DAN~\cite{wang2020few}        & 54.7 & 68.6 & 57.8 & 51.6 &58.2 & 71.9 &- & 57.9 & 69.0 & 60.1 & 54.9 &60.5 & 72.3 &-& - \\ 
            
            & PFENet~\cite{tian2020prior}  & 60.5 & 69.4 & 54.4 & 55.9 &60.1 & 72.9 &- & 62.8 & 70.4 & 54.9 & 57.6 &61.4 & 73.5 &-& 10.8M \\
            & ASGNet~\cite{li2021adaptive}  &59.8 &67.4 &55.6  &54.4 &59.3 &71.7& -&64.6 &71.3 &64.2 &57.3   &64.4 &75.2 &-&10.4M \\
            & CWT~\cite{lu2021simpler}  &56.9 &65.2 &61.2  &48.8 &58.0 &- &-&62.6 &70.2 &\textbf{68.8}  &57.2&64.7  &-&- &- \\
            & RePRI~\cite{boudiaf2021few}  & 59.6 & 68.6 & \underline{62.2} & 47.2 & 59.4 & - &45.1 & 66.2 & 71.4 & 67.0 & 57.7 & 65.6 & - &42.0 & - \\ 
            
            & HSNet~\cite{min2021hypercorrelation}    & 67.3 & 72.3 & 62.0 & \underline{63.1} &\underline{66.2} & \underline{77.6}& \underline{53.9}& 71.8 & \underline{74.4} & 67.0 & \underline{68.3} & \underline{70.4} & \underline{80.6}&\underline{54.4} & \textbf{2.6M} 
            \\
            & CyCTR~\cite{zhang2021few} &\underline{67.2} &\underline{71.1} &57.6  &59.0 &63.7 &73.0&- &\underline{71.0} &75.0 &58.5 &65.0 &67.4 &75.4 &-&- \\\cline{2-17} \\[-2.0ex]
            & VAT (ours)& \textbf{70.0}&\textbf{72.5} &\textbf{64.8}  & \textbf{64.2} & \textbf{67.9} & \textbf{79.6}& \textbf{54.7}& \textbf{75.0} &\textbf{75.2} &\underline{68.4} & \textbf{69.5} & \textbf{72.0} & \textbf{83.2}&\textbf{54.8}&\underline{3.3M} \\

            \bottomrule
    \end{tabular}
    }
    \end{center}\vspace{-5pt}
    \caption{\textbf{Performance comparison on PASCAL-5$^{i}$~\cite{shaban2017one}.} Best results in bold, and second best are underlined.}\label{tab:pascal_sota}\vspace{-10pt}
\end{table}

\begin{table}[t]
    \begin{center}
    \scalebox{0.75}{
    \begin{tabular}{clcccccccccccccc}
                \toprule
                \multirow{2}{*}{\shortstack{Backbone\\feature}} & \multirow{2}{*}{Methods} & \multicolumn{7}{c}{1-shot} & \multicolumn{7}{c}{5-shot} \\ 
                
                & & $20^{0}$ & $20^{1}$ & $20^{2}$ & $20^{3}$ & mean & FB-IoU & mBA& $20^{0}$ & $20^{1}$ & $20^{2}$ & $20^{3}$ & mean & FB-IoU & mBA \\

                \midrule
                
                \multirow{8}{*}{ResNet50~\cite{he2016deep}} & PMM~\cite{yang2020prototype}        & 29.3 & 34.8 & 27.1 & 27.3 & 29.6 & - &-& 33.0 & 40.6 & 30.3 & 33.3 & 34.3 & - &- \\ 
                
                & RPMM~\cite{yang2020prototype}        & 29.5 & 36.8 & 28.9 & 27.0 & 30.6 & - &-& 33.8 & 42.0 & 33.0 & 33.3 & 35.5 & -  &- \\    
                
                & PFENet~\cite{tian2020prior}        & 36.5 & 38.6 & 34.5 & 33.8 & 35.8 & -&- & 36.5 & 43.3 & 37.8 & 38.4 & 39.0 & -&- \\   
                & ASGNet~\cite{li2021adaptive} &- &- &-  &- &34.6 &60.4&- &- &- &- &- &42.5  &67.0 &-\\
                & RePRI~\cite{boudiaf2021few}        & 32.0 & 38.7 &  32.7 & {33.1} & 34.1 & - &6.31& 39.3 & 45.4 & 39.7 & 41.8 & 41.6 & - &4.21\\    
                
                & HSNet~\cite{min2021hypercorrelation}     & 36.3 & \underline{43.1} & 38.7 & 38.7 & 39.2 & \underline{68.2}&\underline{53.0} & \underline{43.3} & \textbf{51.3} & \underline{48.2} & 45.0 & \underline{46.9} & \underline{70.7}&\underline{53.8}\\
                &CyCTR~\cite{zhang2021few} &\underline{38.9} &43.0  &\underline{39.6} &\textbf{39.8} &\underline{40.3} &- &- &41.1 &48.9 &45.2 &\textbf{47.0} &45.6 &-&-\\ \cline{2-16} \\[-2.0ex]
                
                

                
                
              &VAT (ours) &\textbf{39.0} &\textbf{43.8}  &\textbf{42.6} &\underline{39.7} &\textbf{41.3} &\textbf{68.8}&\textbf{54.2}  &\textbf{44.1} &\underline{51.1} &\textbf{50.2} &\underline{46.1} &\textbf{47.9} &\textbf{72.4}&\textbf{54.9}\\

                \bottomrule
        \end{tabular}
        }
    \end{center}\vspace{-5pt}
        \caption{\textbf{Performance comparison on COCO-20$^{i}$~\cite{lin2014microsoft}.}}\label{tab:coco_sota}\vspace{-20pt}
\end{table}

\begin{table}[t]
\begin{center}
\scalebox{0.75}{
        \begin{tabular}{clcccccc}
                \toprule
                \multirow{2}{*}{\shortstack{Backbone\\feature}} & \multirow{2}{*}{Methods} & \multicolumn{2}{c}{mIoU} & \multicolumn{2}{c}{FB-IoU}&\multicolumn{2}{c}{mBA}\\ 
                
                & & 1-shot & 5-shot & 1-shot & 5-shot  & 1-shot & 5-shot \\
                
                \midrule

                \multirow{3}{*}{ResNet50~\cite{he2016deep}}       &FSOT~\cite{liu2021few}&82.5&83.8&-&-&-&-\\
               &HSNet~\cite{min2021hypercorrelation} & \underline{85.5} & \underline{87.8}& \underline{91.0}& \underline{92.5}& \underline{62.1}& \underline{63.3}\\\cline{2-8}\\ [-2.0ex]
                & VAT &\textbf{90.1} &\textbf{90.7} &\textbf{93.8}&\textbf{94.2} &\textbf{68.3}&\textbf{68.4}\\
             
                \midrule
                \multirow{3}{*}{ResNet101~\cite{he2016deep}} & DAN~\cite{wang2020few} & {85.2} & {88.1} &-&-&-&-\\ 
                
                & HSNet~\cite{min2021hypercorrelation}  & \underline{86.5} &\underline{88.5} & \underline{91.6}& \underline{92.9}& \underline{62.4}& \underline{63.6}\\ \cline{2-8} \\[-2.0ex]
                & VAT &\textbf{90.3}  & \textbf{90.8}&\textbf{94.0} & \textbf{94.4}&\textbf{68.0}&\textbf{68.6} \\
                \bottomrule
        \end{tabular}
        }
        \end{center}\vspace{-5pt}
        \caption{\textbf{Mean IoU comparison on FSS-1000~\cite{li2020fss}.}}\label{tab:fss_sota}\vspace{-20pt}
\end{table}

   %
                
  
\section{Experiments}
\subsection{Implementation Details}\label{sec:4.1}
We use ResNet50 and ResNet101~\cite{he2016deep} pre-trained on ImageNet~\cite{deng2009imagenet} and freeze the weights during training, following~\cite{min2021hypercorrelation,zhang2019canet}. No data augmentation is used for training, as explained in the supplementary material. We set the input image sizes to 417 or 473, following~\cite{li2021adaptive,boudiaf2021few}. The window size for Swin Transformer is set to 4. We use AdamW~\cite{loshchilov2017decoupled} with a learning rate of $5\mathrm{e}-4$. Feature maps from conv3$\_$x ($p=3$), conv4$\_$x ($p=4$) and conv5$\_$x ($p=5$) are taken for cost computation. The $K$-shot threshold $\tau$ is set to $0.5$ and the embedding dimension $D$ to $128$. For appearance affinity,  we take the last layers from conv2$\_$x, conv3$\_$x and conv4$\_$x when training on FSS-1000~\cite{li2020fss}, and conv4$\_$x is excluded when training on PASCAL-5$^{i}$~\cite{shaban2017one} and COCO-20$^{i}$~\cite{lin2014microsoft}. We set $c$ to 16, 32, and 64 for conv2$\_$x, conv3$\_$x, and conv4$\_$x.  

\subsection{Experimental Settings}\label{sec:4.2}

\subsubsection{Datasets.}
We evaluate our approach on three standard few-shot segmentation datasets, PASCAL-5$^{i}$~\cite{shaban2017one}, COCO-20$^{i}$~\cite{lin2014microsoft}, and FSS-1000~\cite{li2020fss}. PASCAL-5$^{i}$ contains images from PASCAL VOC 2012~\cite{everingham2010pascal} with added mask annotations~\cite{hariharan2014simultaneous}. It consists of 20 object classes, and as done in OSLSM~\cite{shaban2017one}, they are evenly divided into 4 folds $i \in \{0,1,2,3\}$ for cross-validation, where each fold contains 5 classes. COCO-20$^{i}$ contains 80 object classes, and as done for PASCAL-5$^{i}$, the dataset is evenly divided into 4 folds of 20 classes each. FSS-1000 is a more diverse dataset consisting of 1000 object classes. Following~\cite{li2020fss}, we divide the 1000 categories into 3 splits for training, validation and testing, which consist of 520, 240 and 240 classes, respectively. For PASCAL-5$^{i}$ and COCO-20$^{i}$, we follow the common evaluation practice~\cite{min2021hypercorrelation,tian2020prior,liu2020part} and standard cross-validation protocol, where each fold $i$ is used for evaluation with the other folds used for training. \vspace{-10pt}

\subsubsection{Evaluation Metric.} 
Following common practice~\cite{zhang2019canet,tian2020prior,min2021hypercorrelation,zhang2021few}, we adopt mean intersection over union (mIoU) and foreground-background IoU (FB-IoU) as our evaluation metrics. The mIoU averages over all IoU values for all object classes such that $\mathrm{mIoU} = \frac{1}{C}\sum _{c=1}^{C} \mathrm{IoU}_{c}$, where $C$ is the number of classes in each fold, e.g., $C = 20$ for COCO-20$^{i}$. FB-IoU disregards the object classes and instead averages over foreground and background IoU ($\mathrm{IoU}_\mathrm{F}$ and $\mathrm{IoU}_\mathrm{B}$) such that $\mathrm{FB-IoU} = \frac{1}{2}(\mathrm{IoU}_\mathrm{F} + \mathrm{IoU}_\mathrm{B})$. We additionally adopt Mean Boundary Accuracy (mBA) introduced in~\cite{cheng2020cascadepsp} to evaluate the model's ability to capture fine details. To measure mBA, we first sample 5 radii in $[3, \frac{w+h}{300}]$ at a uniform interval, where $w$ and $h$ are width and height of input image, and average the segmentation accuracy within each radius from the ground-truth boundary.

\subsection{Few-shot Segmentation Results }
Table~\ref{tab:pascal_sota} summarizes quantitative results on PASCAL-5$^{i}$~\cite{shaban2017one}. The tests were conducted on two backbone networks, ResNet50 and ResNet101~\cite{he2016deep}. 
The proposed method outperforms the others on almost all the folds in terms of both mIoU and FB-IoU. It surpasses the others, including HSNet~\cite{min2021hypercorrelation}, in mBA as well, since our ATD helps to improve up-sampling quality by providing higher-level spatial structure for reference. Consistent with this, VAT also attains state-of-the-art performance on COCO-20$^{i}$~\cite{lin2014microsoft}, as shown in Table~\ref{tab:coco_sota}. Interestingly, for the most recent dataset specifically created for few-shot segmentation, FSS-1000~\cite{li2020fss}, VAT outperforms HSNet~\cite{min2021hypercorrelation} and FSOT~\cite{liu2021few} by a large margin, almost a 4.6$\%$ increase in mIoU compared to HSNet with ResNet50 as shown in Table~\ref{tab:fss_sota}. VAT sets a new state-of-the-art for all of these benchmarks. We note that our method outperforms HSNet~\cite{min2021hypercorrelation} despite having more learnable parameters, which is known to have an inverse relation to generalization power~\cite{Tetko1995NeuralNS}, a trend seen in Table~\ref{tab:pascal_sota}. With the proposed method, \textit{i.e.,} 4D convolutional Swin Transformer, that is designed to address the issues like lack of inductive bias, VAT can have a larger number of learnable parameters than that of HSNet~\cite{min2021hypercorrelation}, yet VAT has greater generalization power as well.  

\subsection{Ablation Study}\label{sec:4.4}
We conducted ablations on FSS-1000~\cite{li2020fss}, a large-scale dataset specifically constructed for few-shot segmentation. \vspace{-10pt}

\subsubsection{Effectiveness of each component in VAT.}
\begin{wrapfigure}{r}{0.45\textwidth}
\vspace{-30pt}
  \begin{center}
  \scalebox{0.7}{
  \includegraphics[width=70mm]{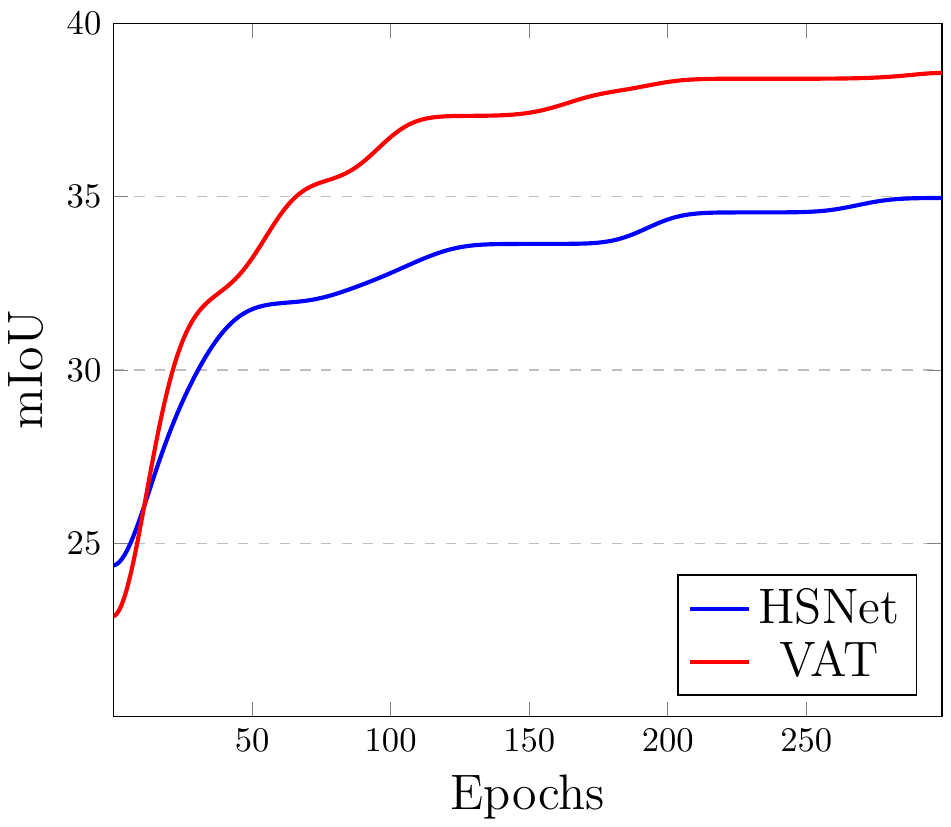}}
  \end{center}\vspace{-15pt}
   \caption{\textbf{Convergence comparison.} Although VAT starts at a lower mIoU, it quickly exceeds HSNet~\cite{min2021hypercorrelation}.}
	\label{fig:comparison}\vspace{-10pt}
\end{wrapfigure}
As the baseline model, we take the architecture composed of VEM and the 2D convolution decoder used in HSNet~\cite{min2021hypercorrelation}. We then progressively add our components one-by-one as shown in Table~\ref{tab:vatable}. Note that we included (\textbf{IV}) and (\textbf{V}) to show the effectiveness of VCM alone and the performance of a model highly similar to HSNet~\cite{min2021hypercorrelation}, respectively. 

As summarized in Table~\ref{tab:vatable}, each component helps to boost performance. Starting from the baseline (\textbf{I}), adding Swin Transformer (\textbf{II}) brings a large gain, which indicates that Swin Transformer effectively performs cost aggregation thanks to its approximated inductive bias and ability to consider spatial structure. When the VEM is replaced by VCM (\textbf{III}), we also observe a significant improvement, which confirms that the issues due to non-overlap are alleviated. We note that (\textbf{IV}) also highlights the importance of inductive bias. As (\textbf{V}) is approximately equivalent to HSNet~\cite{min2021hypercorrelation}, we first compare it with (\textbf{III}), which shows the superiority of the proposed 4D Convolutional Swin Transformer. By including the additional components in (\textbf{VI}) and (\textbf{VII}), the performance is further boosted. Moreover, we observe a large gain in mBA by adding ATD. This shows that the higher-resolution spatial structure provided by appearance embeddings help to refine the fine details.  We additionally provide a visualization of convergence in comparison to HSNet~\cite{min2021hypercorrelation} in Fig.~\ref{fig:comparison}. Thanks to the early convolutions~\cite{xiao2021early}, VAT quickly converges and exceeds HSNet~\cite{min2021hypercorrelation} even though it starts at a lower mIOU.

\vspace{-10pt}

\subsubsection{Base architecture of VTM.}
As summarized in Table~\ref{tab:aggregator}, we provide an ablation study to evaluate the effectiveness of different aggregators for VTM. For cost aggregation, there exists a few learnable aggregators, including MLP-, convolution- and transformer-based aggregators, any of which could be used as a base architecture for VTM. It should be noted that the use of standard transformer~\cite{vaswani2017attention} and MLP-mixer~\cite{tolstikhin2021mlp} is not feasible due to memory requirements. Specifically, we calculated the memory consumption of each and found that using standard transformer requires approximately 84 GB per batch, while the memory for MLP-Mixer could not be measured as it is much greater than standard transformer. Also, we note that the architecture with center-pivot convolutions is equivalent to a deeper version of the architecture with VCM.

\begin{table}[t]
\vspace{-20pt}
    \parbox{.35\linewidth}{
    \begin{center}
    
   \scalebox{0.65}{
   \begin{tabular}{ll|cc|cc}
\toprule
     &\multirow{3}{*}{Components}&\multicolumn{4}{c}{FSS-1000~\cite{li2020fss}} \\
    &&\multicolumn{2}{c}{mIoU ($\%$)}&\multicolumn{2}{c}{mBA ($\%$)} \\
 &&1-shot &5-shot&1-shot &5-shot \\  \midrule
(\textbf{I})&Baseline&80.0&81.8&56.7&56.9 \\
(\textbf{II}) & + Swin Trans.&85.4 &87.4&58.8&59.5 \\
(\textbf{III}) & + VCM&87.0 &88.6 &60.1&61.3 \\
(\textbf{IV}) & only VCM &86.4&88.0&59.6&60.1\\
(\textbf{V}) & (\textbf{IV}) + 4D mix  &86.4 & 87.8&59.9&59.6 \\
(\textbf{VI}) & (\textbf{III}) + GPP  &\underline{87.3} &\underline{88.8} &\underline{60.7}&\underline{61.4}\\
(\textbf{VII}) & + ATD  &\textbf{90.3} &\textbf{90.8}&\textbf{68.0}&\textbf{68.6} \\
\bottomrule
\end{tabular}
}\end{center}\vspace{-5pt}
 \caption{\textbf{Ablation study for VAT.} }\label{tab:vatable}\vspace{-25pt}
    }
    \hfill
    \parbox{.57\linewidth}{
    \centering
    \begin{center}
        
         \scalebox{0.64}{
    \begin{tabular}{l|c|c|c|c}
    \toprule
    \multirow{2}{*}{Different aggregators} & \multicolumn{2}{c}{FSS-1000~\cite{li2020fss}} & Memory & Run-time\\
   &{mIoU (\%)} &mBA (\%)&(GB)&(ms)\\\midrule
    
       Standard transformer~\cite{vaswani2017attention}&OOM &OOM&84 &N/A\\
       MLP-Mixer~\cite{tolstikhin2021mlp}&OOM &OOM&OOM &N/A\\
       Center-pivot 4D convolutions~\cite{min2021hypercorrelation}&\underline{88.1}&\underline{66.5}&\textbf{3.5}&\textbf{52.7}\\
         Linear transformer~\cite{katharopoulos2020transformers}& 87.7&\underline{66.5} &\textbf{3.5}&\underline{56.8}\\
         Fastformer~\cite{wu2021fastformer}& 87.8&66.4 &\textbf{3.5}&122.9\\\midrule
         4D Conv. Swin transformer (Ours) &\textbf{90.3}&\textbf{68.0} &\underline{3.8}&57.3\\
         \bottomrule
    \end{tabular}
    }
  
    \end{center}\vspace{-5pt}
 \caption{\textbf{Ablation study for VTM.} OOM: Out of Memory.}\label{tab:aggregator}\vspace{-25pt}
    }
\end{table}

For a fair comparison, we only replace VTM with another aggregator and leave all the other components in our architecture unchanged. We observe that our method outperforms the other aggregators by a large margin. Interestingly, although center-pivot 4D convolution~\cite{min2021hypercorrelation} also focuses on locality as in Swin Transformer~\cite{liu2021swin}, the performance gap indicates that the ability to adaptively consider pixel-wise interactions is critical. Also, we conjecture that the SW-MSA operation helps to compensate for the lack of global aggregation, which center-pivot convolutions lack. Another interesting point is that Linear Transformer~\cite{katharopoulos2020transformers} and Fastformer~\cite{wu2021fastformer}, which benefit from the global receptive fields of transformers and approximate the self-attention computation, achieve similar performance.   

We additionally provide memory and run-time comparison to other aggregators in Table~\ref{tab:aggregator}. The results are obtained using a single NVIDIA GeForce RTX 3090 GPU and Intel Core i7-10700 CPU. We observe that VAT is relatively slower and consumes more memory. However, 0.3 GB more memory consumption and 5 ms slower run time is a minor sacrifice for better performance. \vspace{-10pt}



\subsubsection{Can VAT also perform well on semantic correspondence?}
To tackle the few-shot segmentation task, we reformulated it as finding semantic correspondences under large intra-class variations and geometric deformations. This suggests that the proposed method could be effective for semantic correspondence as well. Here, we compare VAT to other state-of-the-art methods in semantic correspondence. 

\begin{table}[t]
\begin{center}
\scalebox{0.72}{
\begin{tabular}{l|c|c|c|cccc|cccc|ccc}
        \hlinewd{0.8pt}
        \multirow{3}{*}{Methods}& \multirow{3}{*}{F.T. Feat.}& \multirow{3}{*}{Data Aug.}& \multirow{3}{*}{Cost Aggregation} &\multicolumn{4}{c|}{SPair-71k~\cite{min2019spair}}&\multicolumn{4}{c|}{PF-PASCAL~\cite{ham2017proposal}}&\multicolumn{3}{c}{PF-WILLOW~\cite{ham2016proposal}} \\
        &&&&\multicolumn{4}{c|}{PCK @ $\alpha_{\text{bbox}}$} &\multicolumn{4}{c|}{PCK @ $\alpha_{\text{img}}$}&\multicolumn{3}{c}{PCK @ $\alpha_{\text{bbox}}$}  \\
        &&&&0.03&0.05&0.1&0.15&0.03&0.05&0.1&0.15&0.05&0.1&0.15
        \\
        \midrule
        NC-Net~\cite{Rocco18b}&\cmark&\xmark&4D Conv.& -&- &20.1&- &30.9&54.3&78.9&86.0&33.8&67.0&83.7\\
        SCOT~\cite{liu2020semantic}&- &\xmark&OT-RHM &-&-&35.6&-&-&63.1&85.4&92.7&47.8&76.0&87.1 \\
        CHM~\cite{min2021convolutional}*&\cmark &\xmark&4D Conv.&\underline{14.9}&27.2&46.3&57.5&\underline{67.5}  &\underline{80.1}&{91.6}&94.9&\underline{52.7}&\underline{79.4}& 87.5\\
        MMNet~\cite{zhao2021multi} &\cmark &\xmark&-&-&- &40.9&-&-&77.6&89.1&94.3&-&- &-\\
        PMNC~\cite{Lee_2021_CVPR} &\cmark &\xmark&4D Conv. &-&-&\underline{50.4}&-&\textbf{71.6}&\textbf{82.4}&90.6&-&-&-&- \\\midrule
        \multirow{2}{*}{DHPF~\cite{min2020learning}*} &\cmark &\xmark&RHM&11.0&20.9 &37.3 &47.5&52.0&75.7&90.7&95.0 &49.5&77.6&89.1\\
        &\cmark &\cmark &RHM&-&-&39.4&-&- &-&-&- &-&-&-\\\midrule
        \multirow{2}{*}{CATs~\cite{cho2021semantic}*} &\cmark &\xmark&Transformer &10.2&21.6&43.5&55.0& 41.6&67.5&89.1&94.9 &46.6&75.6&87.5\\
        &\cmark &\cmark&Transformer &13.8&27.7&49.9 &\underline{61.7}&49.9&75.4&\textbf{92.6}&\textbf{96.4} &{50.3}&79.2&90.3\\
    
        \hline
         \multirow{2}{*}{VAT } &\cmark&\xmark &Transformer &\underline{14.9}&\underline{28.3}&48.4&59.1&54.6&72.9&91.1&95.6&46.0&78.8 &\underline{91.3}\\&\cmark
       &\cmark &Transformer&\textbf{19.6}&\textbf{35.0} &\textbf{55.5}&\textbf{65.1}&62.7&78.2&\underline{92.3}&\underline{96.2}&\textbf{52.8}&\textbf{81.6} &\textbf{91.4}\\
        
        \hlinewd{0.8pt}
\end{tabular}
} 
\end{center}\vspace{-5pt}
    \caption{\textbf{Quantitative results on SPair-71k~\cite{min2019spair}, PF-PASCAL~\cite{ham2017proposal} and PF-WILLOW~\cite{ham2016proposal}.} *: The results are obtained using pretrained weights provided by authors or taken from papers.}
    \label{tab:semantic}\vspace{-20pt}
\end{table}
In order to ensure a fair comparison, we note whether each method leverages multi-level features  and  fine-tunes  the  backbone  networks.  We additionally denote the types of cost aggregation. Note that the only difference we made for this experiment is the objective function for loss computation.  Following the common protocol~\cite{min2019hyperpixel,min2020learning,zhao2021multi,huang2019dynamic,min2021convolutional,cho2021semantic}, we use standard benchmarks for this task and our model was trained on the training split of PF-PASCAL~\cite{ham2017proposal} when evaluated on the test split of PF-PASCAL~\cite{ham2017proposal} and PF-WILLOW~\cite{ham2016proposal}, and trained on SPair-71k~\cite{min2019spair} when evaluated on SPair-71k~\cite{min2019spair}. Experimental setting and implementation details can be found in supplementary material.  

As shown in Table~\ref{tab:semantic}, VAT either sets a new state-of-the-art~\cite{min2019spair,ham2016proposal} or attains the second highest PCK~\cite{ham2017proposal}, indicating the importance of cost aggregation in both few-shot segmentation and semantic correspondence.  It also has the potential to benefit general-purpose matching networks as well. Furthermore, when data augmentation is used, we observe a relatively large performance gain compared to DHPF~\cite{min2020learning}, showing that augmentation helps to address the heavy need for data and lack of inductive bias in transformers~\cite{dosovitskiy2020image,cho2021semantic}. Although VAT is on par with state-of-the-art on PF-PASCAL~\cite{ham2017proposal}, we argue that PF-PASCAL~\cite{ham2017proposal} is almost saturated, which makes a comparison difficult. Also, it should be noted that for performance on PF-WILLOW~\cite{ham2016proposal}, VAT outperforms other methods by large margin, which clearly shows superior generalization power of the proposed 4D Convolutional Swin Transformer.

\section{Conclusion}
In this paper, we presented a novel cost aggregation network for few-shot segmentation. To address issues that arise from tokenization of a correlation map for transformer processing, we proposed a 4D Convolutional Swin Transformer, where a high-dimensional Swin Transformer is preceded by a series of small-kernel convolutions. To boost aggregation performance, we applied transformers within a pyramidal structure, and the output is then filtered and in the subsequent decoder with the help of image's appearance embedding. We have shown that the proposed method attains state-of-the-art performance for all the standard benchmarks for both few-shot segmentation and semantic correspondence, where cost aggregation plays a central role. \vspace{-10pt}

\subsubsection{Acknowledgements.}
This research was supported by the MSIT, Korea (IITP-2022-2020-0-01819, ICT Creative Consilience program), and National Research Foundation of Korea (NRF-2021R1C1C1006897). 

\clearpage
%
%
\renewcommand{\thefigure}{\arabic{figure}}

\renewcommand{\thetable}{\arabic{table}}

\setcounter{figure}{0}
\setcounter{table}{0}

\newpage
\begin{center}
	\textbf{\Large Appendix}
\end{center}

In this document, we provide details on the experimental setting, more ablation studies, more quantitative results on semantic correspondence benchmarks, including SPair-71k~\cite{min2019spair}, PF-PASCAL~\cite{ham2017proposal}, and PF-WILLOW~\cite{ham2016proposal}, and more qualitative results on all the benchmarks we used.\vspace{-10pt}

\section*{Appendix A. Experimental Setting for Semantic Correspondence}

\subsubsection{Datasets.}
For the datasets we used, we follow the common protocol~\cite{min2019hyperpixel,min2020learning,Rocco18b,zhao2021multi,huang2019dynamic,min2021convolutional,cho2021semantic} and use standard benchmarks~\cite{ham2016proposal,ham2017proposal,min2019spair}. Specifically, we consider SPair-71k~\cite{min2019spair}, which provides a total of 70,958 image pairs with extreme and diverse viewpoints, scale variations, and rich annotations for each image pair. We also consider relatively small-scale datasets, which include PF-PASCAL~\cite{ham2017proposal} containing 1,351 image pairs from 20 categories and PF-WILLOW~\cite{ham2016proposal} containing 900 image pairs from 4 categories, where each dataset provides corresponding ground-truth annotations.\vspace{-10pt}
\subsubsection{Evaluation metric.}
For evaluation on SPair-71k~\cite{min2019spair}, PF-PASCAL~\cite{ham2017proposal}, and PF-WILLOW~\cite{ham2016proposal}, we employ the percentage of correct keypoints (PCK). It is computed as the ratio of estimated keypoints within the threshold from ground-truths to the total number of keypoints. Concretely, given predicted keypoint $k_\mathrm{pred}$ and ground-truth keypoint $k_\mathrm{GT}$, we count the number of predicted keypoints that satisfy the following condition: $d( k_\mathrm{pred},k_\mathrm{GT}) \leq \alpha \cdot \mathrm{max}(H,W)$, where $d(\,\cdot\,)$ denotes Euclidean distance; $\alpha$ denotes a threshold value; $H$ and $W$ denote height and width of the object bounding box or the entire image, respectively. We evaluate on PF-PASCAL with $\alpha_\mathrm{img}$, and SPair-71k, and PF-WILLOW with $\alpha_\mathrm{bbox}$ following the common protocol. \vspace{-10pt}

\subsubsection{Implementation Details.}
We use ResNet-101~\cite{he2016deep} pre-trained on ImageNet~\cite{deng2009imagenet} for the backbone feature extraction networks. We leave all the components in VAT unchanged. However, we build a different objective function. As in~\cite{min2019hyperpixel,min2020learning,min2021convolutional}, we assume  ground-truth keypoints are provided. We utilize Average End-Point Error (AEPE)~\cite{truong2020glu} and compute it by averaging the Euclidean distance between the ground-truth and estimated flow. Specifically, we compute the loss as $\mathcal{L}=\|F_\mathrm{GT}-F_\mathrm{pred}\|_{2}$, where $F_\mathrm{GT}$ is the ground-truth flow field and $F_\mathrm{pred}$ is the predicted flow field. Note that we achieve this without making any modification to the network architecture. To report the results for different $\alpha$ thresholds, we employ the pre-trained weights released by authors, and simply evaluate without making any changes to their architectures. We use the same data augmentation used in CATs~\cite{cho2021semantic}. For the learning rate, we use the AdamW~\cite{loshchilov2017decoupled} optimizer with $3\mathrm{e}^{-5}$ for VAT and $3\mathrm{e}^{-6}$ for the backbone feature networks. Finally, we use appearance embedding from conv$3\_x$, conv$4\_x$ and conv$5\_x$ as done for FSS-1000~\cite{li2020fss}.

\vspace{-10pt}

\section*{Appendix B. Additional Ablation Study} 

\subsubsection{Ablation study for feature backbone.}
\begin{wraptable}{r}{5.5cm}
\begin{center}
\scalebox{1}{
\begin{tabular}{l|cc}
\toprule
     \multirow{3}{*}{Backbones feature} &\multicolumn{2}{c}{FSS-1000~\cite{li2020fss}} \\
     &\multicolumn{2}{c}{mIoU ($\%$)}\\
&1-shot &5-shot \\  \midrule
ResNet50~\cite{he2016deep} &\underline{90.1} &\underline{90.7}\\
ResNet101~\cite{he2016deep}&\textbf{90.3} &\textbf{90.8}\\
PVT~\cite{wang2021pyramid} &{90.0}&90.6\\
Swin transformer~\cite{liu2021swin} &89.8 &90.2 \\\bottomrule
    \end{tabular}
}
\end{center}
\vspace{-10pt}\caption{\textbf{Ablation study of different feature backbone. }}\label{tab:backbone}
\end{wraptable}
Conventional few-shot segmentation methods only utilized CNN-based feature backbones~\cite{he2016deep} for extracting features.~\cite{zhang2019canet} observed that high-level features contain semantics of objects which could lead to overfitting and is not suitable to use for the task of few-shot segmentation. Then the question naturally arises, what about other networks? As addressed in many works~\cite{raghu2021vision,dosovitskiy2020image}, CNN and transformers see images differently, which means that the kinds of backbone networks may affect the performance significantly, but this has never been explored for this task. We thus exploit several well-known vision transformer architectures to explore the potential differences that may exist. 

The results are summarized in Table~\ref{tab:backbone}. We find that both convolution- and transformer-based backbone networks attain similar performance. We conjecture that although it has been widely studied that convolutions and transformers see differently~\cite{raghu2021vision}, as they are pre-trained on the same dataset~\cite{deng2009imagenet}, the representations learned by models are almost alike.  Note that we only utilized backbones with a pyramidal structure, and the results may differ if other backbone networks are used, which we leave for future exploration. \vspace{-10pt}

\subsubsection{Effectiveness of Data Augmentation. }
We explore the effectiveness of data augmentation for few-shot segmentation. In this experiment, we employ two types of data augmentation, which are introduced either in PFE-Net~\cite{tian2020prior} or CATs~\cite{cho2021semantic}. We summarize the augmentation types in Table~\ref{pfeaug} and Table~\ref{catsaug}. For this ablation study, we use two datasets, PASCAL-5$^i$~\cite{shaban2017one} and FSS-1000~\cite{li2020fss}.  The results are summarized in Table~\ref{ablation}. Note that we use the same augmentation types and probability as theirs. For a fair comparison, we keep all the other experimental settings the same, \textit{e.g.,} number of iterations and learning rate.  

\begin{table}[]
    \centering
    \scalebox{1}{
   \begin{tabular}{c|c|ccccc|c}
\toprule
     \multirow{3}{*}{PFE-Net Aug.~\cite{tian2020prior}} &\multirow{3}{*}{CATs Aug.~\cite{cho2021semantic}} &\multicolumn{5}{c|}{PASCAL-5$^i$~\cite{shaban2017one}} &{FSS-1000~\cite{li2020fss}} \\
     & &\multicolumn{5}{c|}{mIoU ($\%$)}&{mIoU ($\%$)}\\
     && $5^{0}$ & $5^{1}$ & $5^{2}$ & $5^{3}$ &mean &\\
   \midrule
   \xmark &\xmark &\textbf{70.0} &\textbf{72.5} &\textbf{64.8} &\underline{64.2} &\textbf{67.9} &\textbf{90.3}\\
   \cmark &\xmark &\underline{68.4} &\underline{72.3} &64.4 &63.9 &\underline{67.3} &90.0\\
   \xmark &\cmark &65.7 &72.2 &62.3 &\textbf{64.3} &66.1 &90.1\\
   \cmark &\cmark &65.2 &71.1 &63.2 &63.4 &65.7 &\underline{90.2}\\\bottomrule

    \end{tabular}}
        \caption{\textbf{Ablation study of Data Augmentation. }}\label{ablation}
\end{table}

As PFE-Net~\cite{tian2020prior} does not address the effectiveness of data augmentation and CATs~\cite{cho2021semantic} is designed for the semantic correspondence task, we are the first to analyze the effectiveness of data augmentation in the few-shot segmentation setting. Overall, we observe that using the data augmentation techniques severely affects the overall performance. Interestingly, although the augmentation technique introduced by CATs~\cite{cho2021semantic} showed a significant performance boost in the semantic correspondence task, it attains the lowest mIoU when evaluated on PASCAL-5$^i$~\cite{shaban2017one} and the second lowest for FSS-1000~\cite{li2020fss}. The severe performance drop in PASCAL-5$^i$~\cite{shaban2017one} indicates a detrimental influence of using CATs~\cite{cho2021semantic} data augmentation. However, given the small difference to the best performance (0.3$\%$) for FSS-1000~\cite{li2020fss}, the results may differ in a retrial. For PFE-Net~\cite{tian2020prior} data augmentation, we observe results to be on par with the best reported results. However, at fold 0, there is a large gap between them, which indicates the detrimental effects of data augmentation on performance. Using both augmentations results in a large performance drop for PASCAL-5$^i$~\cite{shaban2017one}, arguably due to the detrimental effects of both augmentations, but for FSS-1000~\cite{li2020fss}, we observe only a small difference. 

\begin{table*}
    \parbox{.5\linewidth}{
    \centering
         \begin{tabular}{cl|c}
       \toprule
        &Augmentation type &Probability \\
        \midrule
        \textbf{(I)} &ToGray  &0.2 \\
        \textbf{(II)} &Posterize &0.2 \\
        \textbf{(III)} &Equalize &0.2 \\
        \textbf{(IV)} &Sharpen &0.2 \\
        \textbf{(V)} &RandomBrightnessContrast &0.2 \\
        \textbf{(VI)} &Solarize &0.2 \\
        \textbf{(VII)} &ColorJitter &0.2 \\\bottomrule

\end{tabular}
    \caption{\textbf{CATs~\cite{cho2021semantic} Aug. Type. }}
    \label{catsaug}
       
    }
    \hfill
    \parbox{.45\linewidth}{
    \centering
        
       \begin{tabular}{cl|c}
       \toprule
        &Strong Aug. type &Probability \\
        \midrule
        \textbf{(I)} &RandScale  &1 \\
        \textbf{(II)} &Crop &1 \\
        \textbf{(III)} &Gaussian Blur &0.5 \\
        \textbf{(IV)} &Horizontal Flip &0.5 \\
         \textbf{(V)} &Rotate &0.5 \\
        \bottomrule

\end{tabular}
    \caption{\textbf{PFE-Net~\cite{tian2020prior} Aug. Type.}}
    \label{pfeaug}
   
    }
\end{table*}
Consequently, we conjecture that the detrimental effects on PASCAL-5$^i$~\cite{shaban2017one} and seemingly trivial effects on FSS-1000~\cite{li2020fss} could be attributed to a few reasons: First, as shown in Table~\ref{ablation}, since the difference between the results of the non-data augmentation approach and the PFE-Net~\cite{tian2020prior} augmentation approach is only 0.6$\%$ for PASCAL-5$^i$, this may be due to the implementation details. For the training, we followed HSNet~\cite{min2021hypercorrelation} to force randomness for diverse episode combinations, which may have made such a gap. Second, although the data augmentation may help transformers by providing inductive bias and addressing the heavy need for data, for few-shot setting, where the objective is to predict labels of unseen classes, the results may be different to that of semantic correspondence. It was demonstrated~\cite{cho2021semantic} that for semantic correspondence, data augmentation indeed helps to boost the performance, but a different problem formulation for few-shot segmentation may result in detrimental effects. Third, since we act on correlation maps, applying data augmentation may significantly affect the matching distribution at each pixel. Unlike those works directly working on feature refinement~\cite{tian2020prior,zhang2021few}, where adopting data augmentation has a direct influence on feature maps, VAT aggregates the correlation maps computed between the features extracted from augmented images, which may result in different effects (performance drop) when the objective is to predict unseen classes. Lastly, combining both augmentations may increase the difficulty of learning, which in turn impacts accuracy.\vspace{-10pt}

\subsubsection{Ablation study for ATD.}
In this ablation study, we show a quantitative comparison between the proposed ATD and a decoder without transformers~\cite{vaswani2017attention,wang2020linformer,wu2021fastformer,liu2021swin} to find out whether the model benefits from the use of transformers for further cost aggregation and filtering with the aid of the appearance embedding. For convenience, we call this Appearance-aware Decoder (AD). To implement this, we only exclude the transformers within ATD and leave all the other components and training settings unchanged, \textit{e.g.,} network architecture, hyperparameters, learning rate and number of iterations.\vspace{-10pt}
\begin{table}[]
    \centering
   \begin{tabular}{l|cc|cc|cc}
\toprule
     \multirow{3}{*}{Components} &\multicolumn{6}{c}{FSS-1000~\cite{li2020fss}} \\
     &\multicolumn{2}{c}{mIoU ($\%$)}&\multicolumn{2}{c}{FB-IoU ($\%$)}&\multicolumn{2}{c}{mBA ($\%$)}\\
 &1-shot &5-shot &1-shot &5-shot &1-shot &5-shot\\  \midrule
Convolutions& 87.3&88.8&92.2&93.2&66.8&67.2\\
Transformers&90.3&90.8&94.0&94.4&68.0&68.6\\\bottomrule
    \end{tabular}
        \caption{\textbf{Ablation study for ATD. }}\label{tab:decoder}
\end{table}
As shown in Table~\ref{tab:decoder}, we observe a large performance gap between AD and ATD, which demonstrates that using transformer allows for more effective aggregation, filtering and integration of correlation maps and appearance embedding. More specifically, we observe a 3$\%$ mIoU difference and find similar differences for FB-IoU and mBA. Without using transformers, where only convolutions are used, we observe that the results are equal  to that of (\textbf{VI}) in the ablation study for VAT. This indicates that meaningful aggregation may not have occurred. It should be noted that we observe highly competitive results for mBA for both approaches, confirming a positive effect from the high-resolution spatial structure of the appearance embedding.\vspace{-10pt}  

\subsubsection{Ablation study for VCM.}
For this ablation study, we aim to further support our claims that the VCM (overlapping convolutions) compensates for the lack of inductive bias and alleviates the detrimental effects caused at window boundaries. To this end, we use Linear transformer~\cite{katharopoulos2020transformers}, Fastformer~\cite{wu2021fastformer} and Swin Transformer~\cite{liu2021swin} to validate the effectiveness. Note that we already reported the results for the ones with VCM, but we additionally provide FB-IoU and mBA results. For the implementation of VEM, we refer the readers to Algorithm 1.

\begin{table}[]
    \centering
   \begin{tabular}{l|cc|cc|cc}
\toprule
     \multirow{3}{*}{Components} &\multicolumn{6}{c}{FSS-1000~\cite{li2020fss}} \\
     &\multicolumn{2}{c}{mIoU ($\%$)}&\multicolumn{2}{c}{FB-IoU ($\%$)}&\multicolumn{2}{c}{mBA ($\%$)}\\
 &1-shot &5-shot &1-shot &5-shot &1-shot &5-shot\\  \midrule
VEM + Linear Transformer~\cite{katharopoulos2020transformers}&87.0 &87.4&90.7&91.0&65.0&64.9\\
VEM + Fastformer~\cite{wu2021fastformer}&87.1&87.6&90.9&91.2&65.3&65.2\\
VEM + Swin Transformer~\cite{liu2021swin}&89.9&90.5&92.9&94.0&67.8&68.2\\\midrule
VCM + Linear Transformer~\cite{katharopoulos2020transformers}& 87.7&88.3&92.3&92.2&66.5&66.7\\
VCM + Fastformer~\cite{wu2021fastformer}& 87.8&88.2&91.8&91.9&66.4&66.4\\
VCM + Swin Transformer~\cite{liu2021swin}&90.3&90.8&94.0&94.4&68.0&68.6\\

\bottomrule
    \end{tabular}
        \caption{\textbf{Ablation study for VCM. }}\label{vcm}
\end{table}

As shown in Table~\ref{vcm}, we find a similar pattern to the results for VCM. Swin Transformer attained the best results, while Linear Transformer~\cite{katharopoulos2020transformers} and Fastformer~\cite{wu2021fastformer} show similar results. Interestingly, when VCM is replaced with VEM, the performance difference for Swin Transformer and the other two differ substantially. Specifically, for Swin Transformer, the mIoU is 89.9$\%$ when equipped with VEM, which is a 0.4$\%$ performance drop and is a relatively lower drop compared to those of Linear Transformer and Fastformer. This could be due to the relative position bias that Swin Transformer provides, which the other two transformers lack. Furthermore, we suspect that the lower mIoU results could be explained by one of the following factors: simplified self-attention computation, local smoothness property of a correlation map, and consideration of spatial structure.  \vspace{-10pt}

\section*{Appendix C. Limitations}
\begin{figure}[t]
\centering
\includegraphics[width=0.5\textwidth]{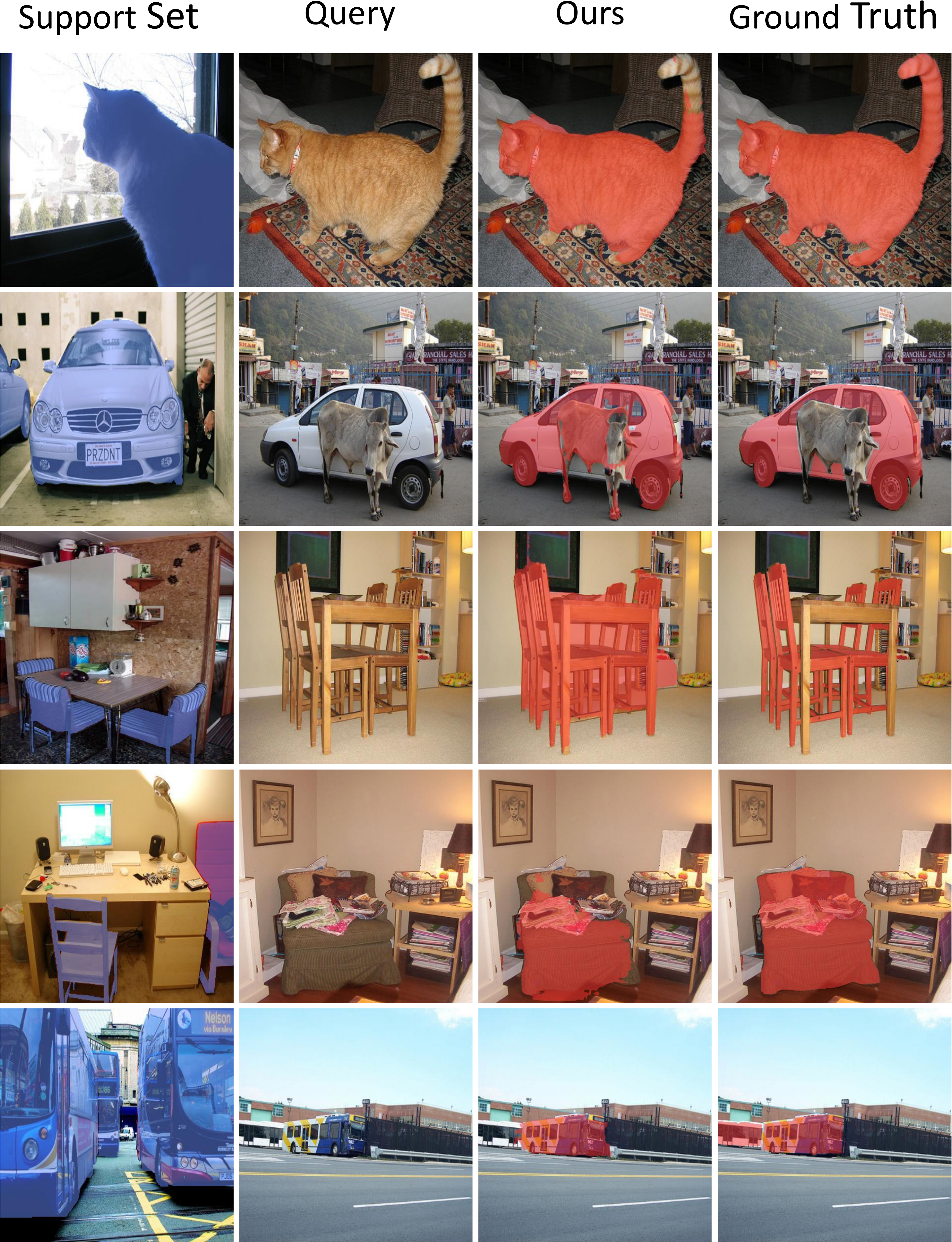}
\vspace{-10pt}\caption{\textbf{Failure cases.}   }
\label{failure}
\end{figure}

An apparent limitation is that since our approach acts on correlation maps, we need to explicitly compute the global correlation maps and store them. This is indeed memory expensive, and increases with the spatial resolution of the correlation maps. Although we utilize a coarse-to-fine architecture, this does not make the training feasible when resolutions are high. Specifically, given a spatial resolution of feature maps at size 128$\times$128, the resultant size of correlation maps is at least 128$^4$, and counting the level dimensions as well as other pyramidal levels $p$, it is difficult to train with a sufficient batch size even with NVIDIA GeForce RTX-3090 GPUs. This might limit the accessibility of this approach. We also visualize failure cases in Fig.~\ref{failure}. \vspace{-10pt}

\clearpage

\section*{Appendix D. More Results}

\vspace{-10pt}
\subsubsection{Quantitative Results for Semantic Correspondence.}
As shown in Table~\ref{tab:spair}, we provide per-class quantitative results on  SPair-71k~\cite{min2019spair} in comparison to other semantic correspondence methods, including CNNGeo~\cite{rocco2017convolutional}, WeakAlign~\cite{rocco2018end}, NC-Net~\cite{Rocco18b}, HPF~\cite{min2019hyperpixel}, SFNet~\cite{lee2019sfnet}, DCC-Net~\cite{huang2019dynamic}, GSF~\cite{jeon2020guided}, SCOT~\cite{liu2020semantic}, DHPF~\cite{min2020learning}, CHM~\cite{min2021convolutional}, MMNet~\cite{zhao2021multi}, PMNC~\cite{Lee_2021_CVPR} and CATs~\cite{cho2021semantic}. \vspace{-10pt}

\begin{table*}[]
    \begin{center}
        \scalebox{0.7}{
        \begin{tabular}{l|cccccccccccccccccc|c}
        \toprule
         Methods & aero. & bike & bird & boat & bott. & bus & car & cat & chai. & cow & dog & hors. & mbik. & pers. & plan. & shee. & trai. & tv & all\\
        \midrule

        CNNGeo~\cite{rocco2017convolutional} &  23.4 & 16.7 & 40.2 & 14.3 & 36.4 & 27.7 & 26.0 & 32.7 & 12.7 & 27.4 & 22.8 & 13.7 & 20.9 & 21.0 & 17.5 & 10.2 & 30.8 & 34.1 & 20.6  \\
      
       WeakAlign~\cite{rocco2018end} &  22.2 & 17.6 & 41.9 & {15.1} & {38.1} & {27.4} & {27.2} & 31.8 & 12.8 & 26.8 & 22.6 & 14.2 & 20.0 & 22.2 & 17.9 & 10.4 & {32.2} & 35.1 & 20.9 \\
    
       NC-Net~\cite{Rocco18b} & 17.9 & 12.2 & 32.1 & 11.7 & 29.0 & 19.9 & 16.1 & 39.2 & 9.9 & 23.9 & 18.8 & 15.7 & 17.4 & 15.9 & 14.8 & 9.6 & 24.2 & 31.1 & 20.1   \\\\[-2.3ex]

        HPF~\cite{min2019hyperpixel} & {25.2} & {18.9} & {52.1} & {15.7} & {38.0} & {22.8} & {19.1} & {52.9} & {17.9} & {33.0} & {32.8} & {20.6} & {24.4} & {27.9} & 21.1 & {15.9} & {31.5} & {35.6} & {28.2} \\\\[-2.3ex]
        
        {SCOT}~\cite{liu2020semantic} & {34.9} & {20.7} & {63.8} & {21.1} & {43.5} & {27.3} & {21.3} & {63.1} & {20.0} & {42.9} & {42.5} & {31.1} & {29.8} & {35.0} & {27.7} & {24.4} & {48.4} & {40.8} & {35.6} \\
       
       {DHPF}~\cite{min2020learning} & {38.4} & {23.8} & {68.3} & {18.9} & {42.6} & {27.9} & {20.1} & {61.6} & {22.0} & {46.9} & {46.1} & {33.5} & {27.6} & {40.1} & {27.6} & {28.1} & {49.5} & {46.5} & {37.3} \\

        {CHM}~\cite{min2021convolutional} & {49.1} & {33.6} & {64.5} & {32.7} & {44.6} & {47.5} & {43.5} & {57.8} & {21.0} & {61.3} & {54.6} & {43.8} & {35.1} & {43.7} & {38.1} & {33.5} & {70.6} & {55.9} & {46.3} \\
        MMNet~\cite{zhao2021multi} &43.5&27.0&62.4&27.3&40.1&50.1&37.5&60.0&21.0&56.3&50.3&41.3&30.9&19.2&30.1&33.2&642&43.6&40.9  \\
        PMNC~\cite{Lee_2021_CVPR}  &\underline{54.1}  &35.9  &\underline{74.9}  &\underline{36.5} &42.1  &48.8  &40.0 &\textbf{72.6}  &21.1  &{67.6}  &\underline{58.1}  &50.5  &40.1  &\textbf{54.1}  &\underline{43.3}  &35.7  &{74.5}  &59.9  &\underline{50.4}  \\
        CATs~\cite{cho2021semantic}  & {52.0} & {34.7} & {72.2} & {34.3} &\underline{49.9} & \underline{57.5} & \underline{43.6} & {66.5} & \textbf{24.4} & {63.2} & {56.5} & \underline{52.0} & \underline{42.6} & {41.7} & {43.0} & {33.6} & {72.6} & {58.0} & {49.9} \\\midrule
        VAT$\dagger$ (ours)  &{49.8}& \underline{36.8}& {70.1}& {33.5} &{46.1} &{46.0}   &{31.1} &\underline{69.9}  &{15.7}  &\underline{69.9}  &{57.2}  &{47.2}  &{38.5}  &{41.8}  &{43.0}  &\underline{35.5}  &\underline{75.0} &\underline{61.8} &{48.4}  \\
        VAT (ours)  &\textbf{58.8}& \textbf{40.0}& \textbf{75.3}& \textbf{40.1} &\textbf{52.1} &\textbf{59.7}   &\underline{44.2} &69.1  &\underline{23.3}  &\textbf{75.1}  &\textbf{61.9}  &\textbf{57.1}  &\underline{46.4}  &\underline{49.1}  &\textbf{51.8}  &\textbf{41.8}  &\textbf{80.9}&\textbf{70.1}  &\textbf{55.5}  \\
       \bottomrule
        \end{tabular}} 
    \caption{\label{tab:spair} \textbf{Per-class quantitative evaluation on SPair-71k~\cite{min2019spair} benchmark.}}\vspace{-10pt}
    
    \end{center}\vspace{-10pt}
\end{table*}
\begin{table}[t]
    \begin{center}
    \scalebox{1}{
    \begin{tabular}{cl|ccccc|ccccc}
            \toprule
            \multirow{2}{*}{\shortstack{Backbone\\network}} & \multirow{2}{*}{Methods} & \multicolumn{5}{c|}{1-shot} & \multicolumn{5}{c}{5-shot}  \\ 
            
            & & $5^{0}$ & $5^{1}$ & $5^{2}$ & $5^{3}$ &mean& $5^{0}$ & $5^{1}$ & $5^{2}$ & $5^{3}$ &mean  \\
            \midrule
             \multirow{3}{*}{ResNet50~\cite{he2016deep}} 
                & RePRI~\cite{boudiaf2021few}        &45.8 &53.7  &46.6   &50.0   &49.0&45.4  &46.9  &41.8  &41.0   &43.8\\    
                
                & HSNet~\cite{min2021hypercorrelation}     &53.9  &54.7  &53.3 &53.6 &53.9 &54.6&55.1  &54.0  &54.2  &54.5\\
            
                
                

                
                
              &VAT (ours) &55.1 &55.1  &53.8 &53.6 &54.4  &55.4 &55.3 &54.5 &53.9 &54.8\\

            \midrule
            
             \multirow{3}{*}{ResNet101~\cite{he2016deep}} 
                & RePRI~\cite{boudiaf2021few}        &47.6 &47.6  &41.9   &43.3   &45.1&46.4  &44.4  &38.4  &38.7   &42.0\\    
                
                & HSNet~\cite{min2021hypercorrelation}     & 53.9 & 54.4 &53.5 &53.9 &53.9 &54.3&54.7  &54.2  &54.2  &54.4\\
            
                
                

                
                
              &VAT (ours) &54.7 &54.6  &53.9 &55.5 &54.7  &55.0 &55.0 &54.5 &54.8 &54.8\\

            \bottomrule
    \end{tabular}
    }
    \end{center}\vspace{-10pt}
    \caption{\textbf{mBA comparison on PASCAL-5$^{i}$~\cite{shaban2017one}.} }\label{tab:pascalmba}\vspace{-20pt}
\end{table}

\begin{table}[t]
    \begin{center}
    \scalebox{1}{
    \begin{tabular}{clcccccccccccc}
                \toprule
                \multirow{2}{*}{\shortstack{Backbone\\feature}} & \multirow{2}{*}{Methods} & \multicolumn{5}{c}{1-shot} & \multicolumn{5}{c}{5-shot} \\ 
                
                & & $20^{0}$ & $20^{1}$ & $20^{2}$ & $20^{3}$ & mean & $20^{0}$ & $20^{1}$ & $20^{2}$ & $20^{3}$ & mean \\

                \midrule
                
                \multirow{3}{*}{ResNet50~\cite{he2016deep}} 
                & RePRI~\cite{boudiaf2021few}        &6.84 &6.16  &5.76   &6.46   &6.31&5.44  &4.45  &3.49  &3.47   &4.21\\    
                
                & HSNet~\cite{min2021hypercorrelation}     &53.1  &52.9  & 53.0&53.0 &53.0 &53.6&53.8  &54.1  &53.7  &53.8\\
            
                
                

                
                
              &VAT (ours) &54.1 &54.0  &54.5 &54.0 &54.2  &54.6 &54.8 &55.4 &54.7 &54.9\\

                \bottomrule
        \end{tabular}
        }
    \end{center}\vspace{-10pt}
        \caption{\textbf{mBA comparison on COCO-20$^{i}$~\cite{lin2014microsoft}.}}\label{tab:cocomba}\vspace{-10pt}
\end{table}

\subsubsection{More results for mBA comparison.}
In Table~\ref{tab:pascalmba} and Table~\ref{tab:cocomba}, we provide per fold quantitative results for mBA. Note that we obtained the mBA results for HSNet~\cite{min2021hypercorrelation} and RePRI~\cite{boudiaf2021few} using the pre-trained weights and code released by the authors. We omit the results for   CyCTR~\cite{zhang2021few} as the official code and weights by the authors are not publicly available. 

\subsubsection{Qualitative Results.}
\begin{figure*}[t]
\centering
\includegraphics[width=0.99\textwidth]{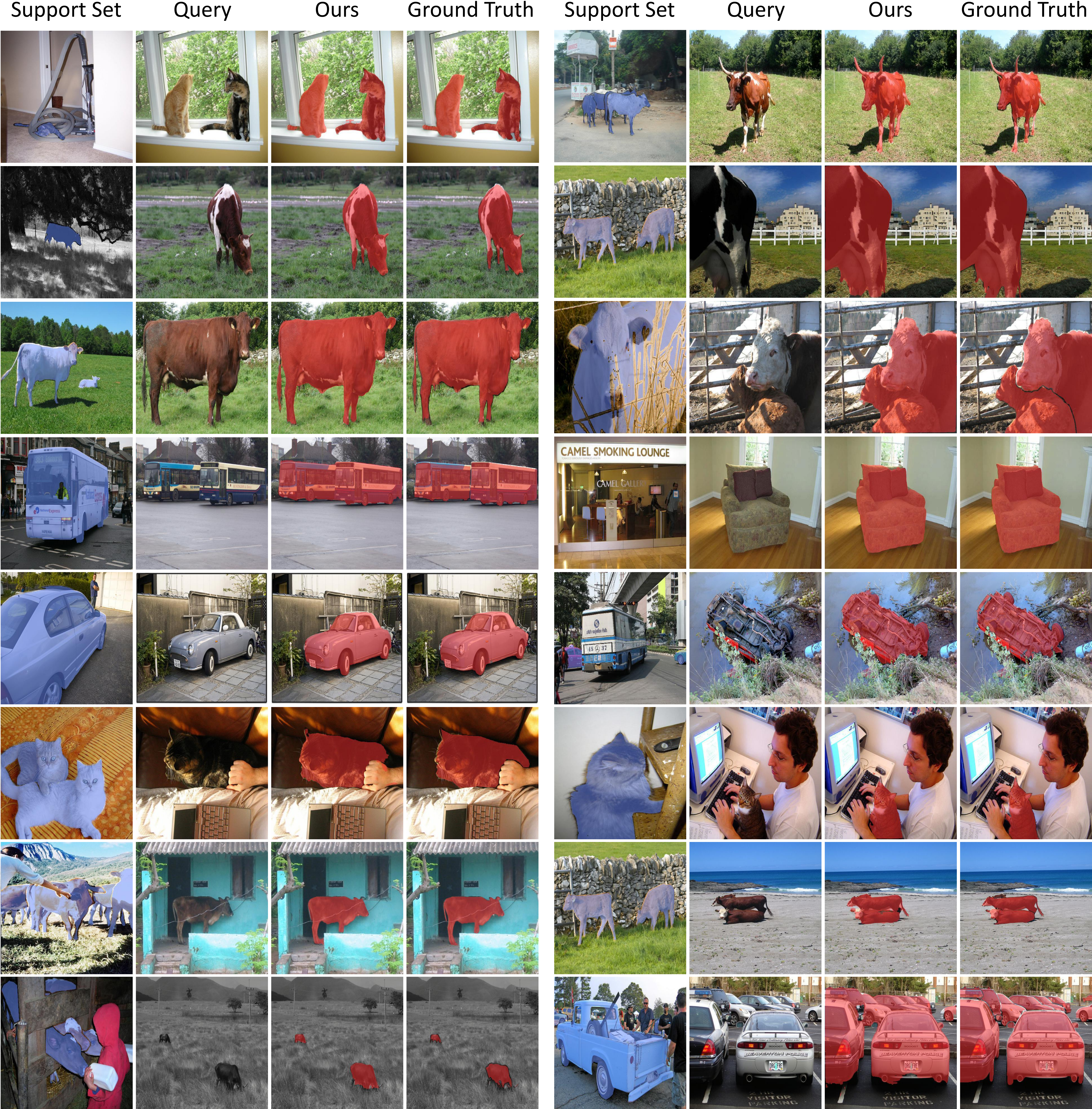}\vspace{-10pt}
\caption{\textbf{Qualitative results on PASCAL-5$^{i}$~\cite{shaban2017one}.}   }
\label{PASCAL}
\end{figure*}

\begin{figure*}[t]
\centering
\includegraphics[width=0.99\textwidth]{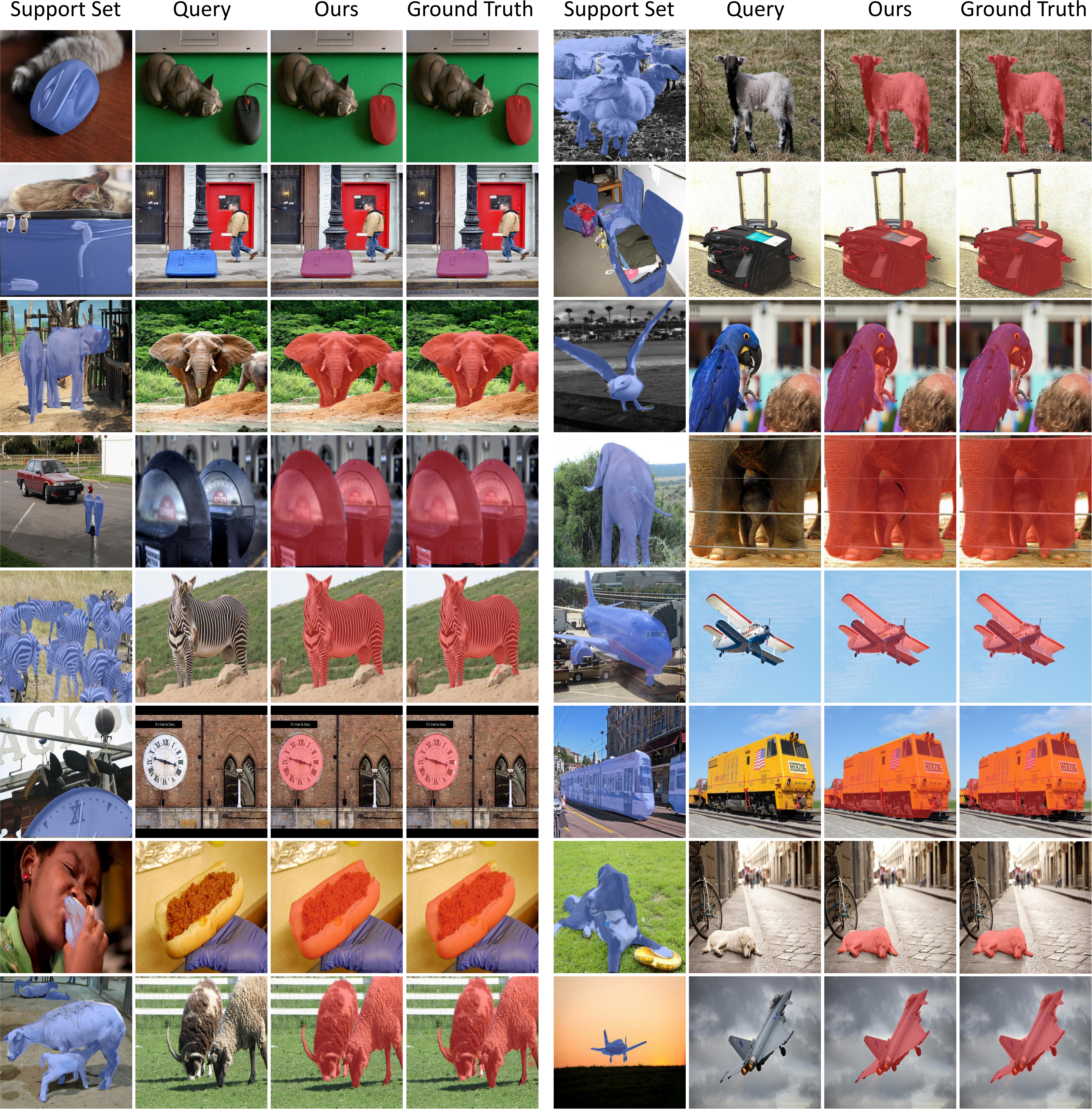}\vspace{-10pt}
\caption{\textbf{Qualitative results on COCO-20$^{i}$~\cite{lin2014microsoft}.}   }
\label{COCO}
\end{figure*}

\begin{figure*}[t]
\centering
\includegraphics[width=0.99\textwidth]{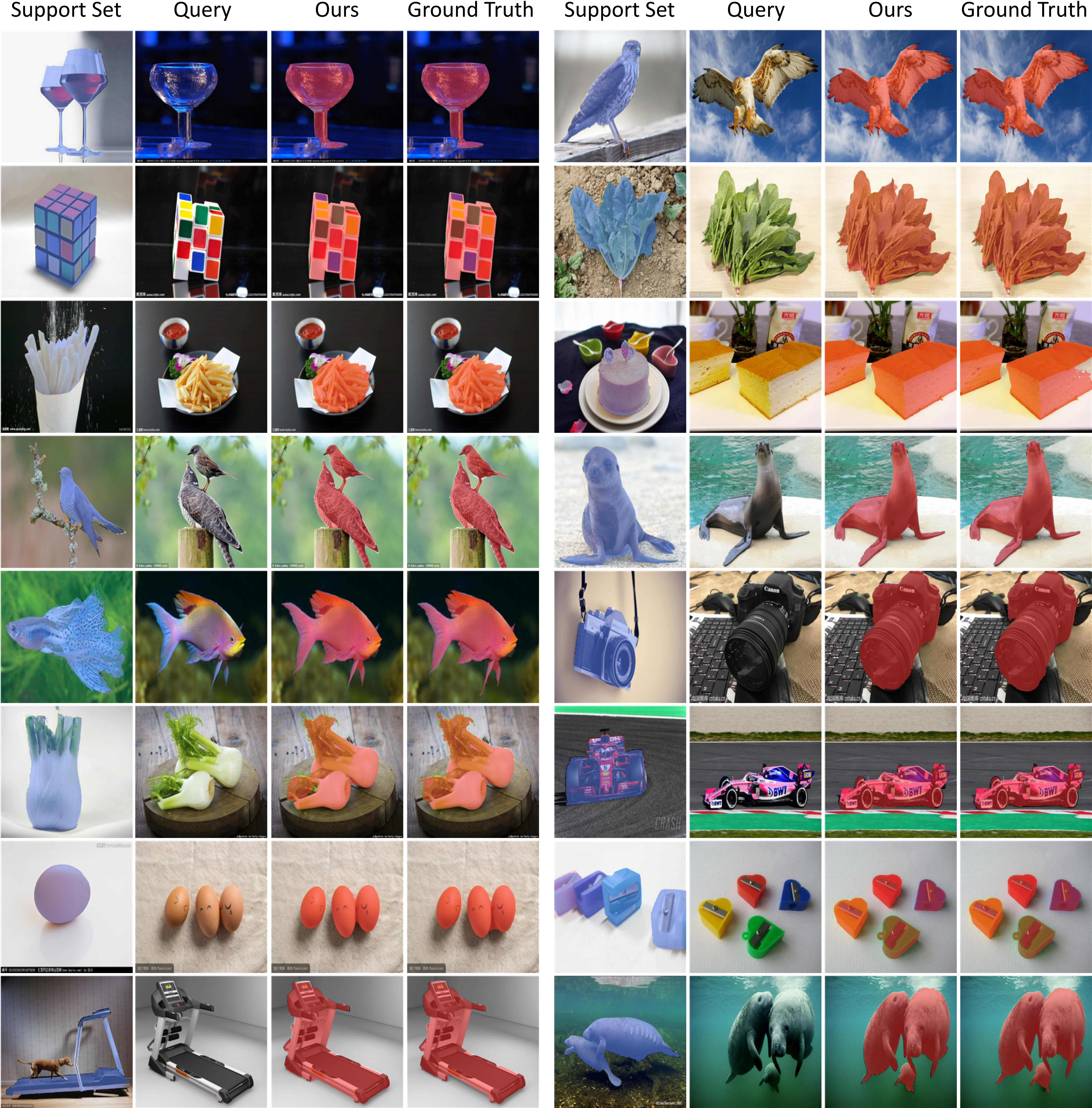}\vspace{-10pt}
\caption{\textbf{Qualitative results on FSS-1000~\cite{li2020fss}.}   }
\label{FSS}
\end{figure*}

\begin{figure*}[t]
\centering
\includegraphics[width=0.99\textwidth]{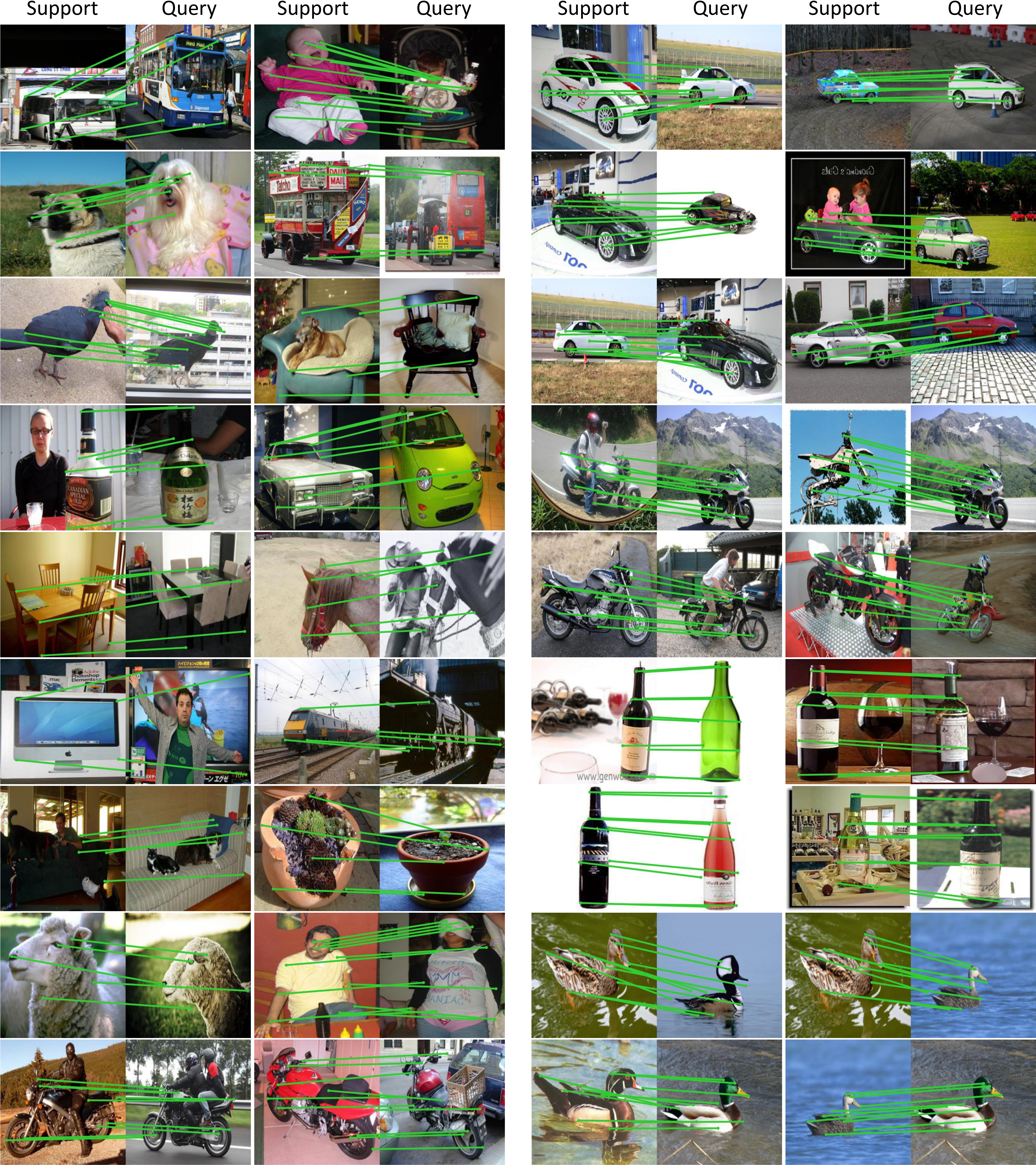}\vspace{-10pt}
\caption{\textbf{Qualitative results on PF-PASCAL~\cite{ham2017proposal} (left) and PF-WILLOW~\cite{ham2016proposal} (right). }   }
\label{WILLOW}
\end{figure*}

\begin{figure*}[t]
\centering
\includegraphics[width=0.8\textwidth]{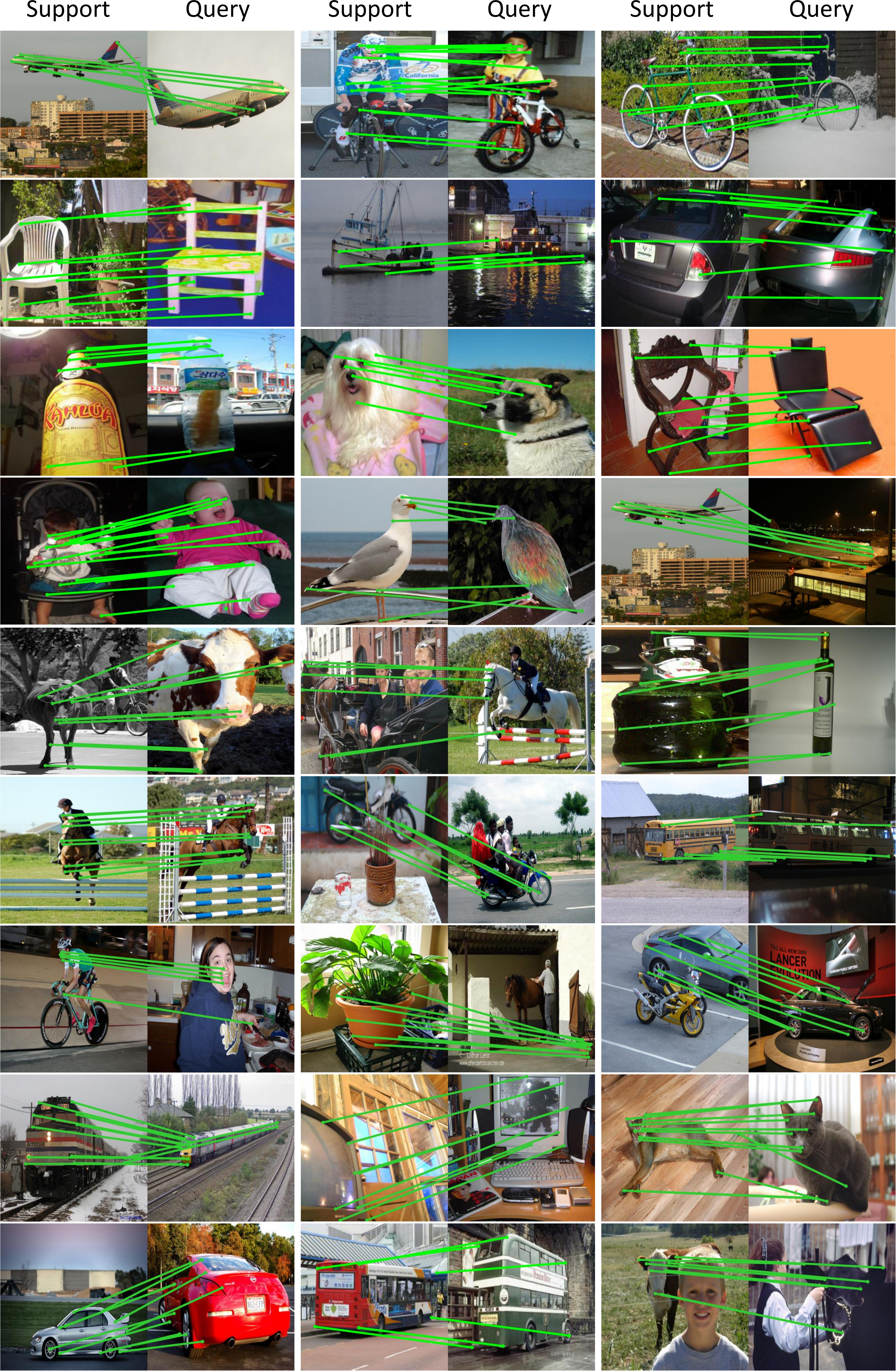}\vspace{-10pt}
\caption{\textbf{Qualitative results on SPair-71k~\cite{min2019spair}.}   }
\label{SPAIR}
\end{figure*}

As shown in Figure~\ref{PASCAL}, Figure~\ref{COCO}, Figure~\ref{FSS}, Figure~\ref{WILLOW} and Figure~\ref{SPAIR}, we provide qualitative results on all the benchmarks, which includes PASCAL-5$^{i}$~\cite{shaban2017one}, COCO-20$^{i}$~\cite{lin2014microsoft}, FSS-1000~\cite{li2020fss}, PF-PASCAL~\cite{ham2017proposal}, PF-WILLOW~\cite{ham2016proposal} and SPair-71k~\cite{min2019spair}.
\newpage

\clearpage

\clearpage
\clearpage
\clearpage
\clearpage
\clearpage
\clearpage

\bibliographystyle{splncs04}
\bibliography{egbib}

\end{document}